\documentclass[10pt]{IEEEtran}
\usepackage{amsmath,amsfonts}
\usepackage{etoolbox}
\usepackage{array}
\usepackage{pifont}
\usepackage[caption=false,font=normalsize,labelfont=sf,textfont=sf]{subfig}
\usepackage{textcomp}
\usepackage{stfloats}
\usepackage{url}
\usepackage{verbatim}
\usepackage{graphicx}
\usepackage{cite}
\usepackage{float}
\usepackage{adjustbox}
\usepackage{caption}
\usepackage[breaklinks,colorlinks,colorlinks=true]{hyperref}
\usepackage[capitalize]{cleveref}

\usepackage{booktabs,makecell, multirow, tabularx, tabularray}
\usepackage{latexsym}
\usepackage{algorithmic,algorithm}
\usepackage{subcaption}
\usepackage{graphbox}
\usepackage{xspace}

\makeatletter
\DeclareRobustCommand\onedot{\futurelet\@let@token\@onedot}
\def\@onedot{\ifx\@let@token.\else.\null\fi\xspace}
\def\eg{\emph{e.g}\onedot}

\def\etal{\emph{et al}\onedot}
\def\figureautorefname{Fig.}

\crefformat{figure}{#2\figureautorefname~#1#3}
\def\tableautorefname{Tab.}
\crefformat{table}{#2\tableautorefname~#1#3}
\makeatother
\begin{document}

\title{Prompt-Driven Building Footprint Extraction in Aerial Images with Offset-Building Model}

\author{
        Kai Li,\thanks{* Jingbo Chen is the corresponding author.}
        \thanks{Kai Li, Junxian Ma, and Chenhao Wang are with Aerospace Information Research Institute, Chinese Academy of Sciences, Beijing 100101, China, and also with School of Electronic, Electrical and Communication Engineering, University of Chinese Academy of Sciences, Beijing 100049, China. \url{likai211@mails.ucas.ac.cn}}
        \thanks{Kai Li is also a Joint PhD student in Applied Machine Learning Lab, School of Data Science, City University of Hong Kong, Kowloon Tong, Hong Kong 999077}
        Yupeng Deng, 
        Yunlong Kong, 
        Diyou Liu,
        Jingbo Chen\textsuperscript{*},
        Yu Meng,\thanks{Yupeng Deng, Yunlong Kong, Diyou Liu, Jingbo Chen, and Yu Meng are with Aerospace Information Research Institute, Chinese Academy of Sciences, Beijing 100101, China (e-mail: \{dengyp, kongyl, liudy, chenjb, mengyu\}@aircas.ac.cn)}
        Junxian Ma,
        and Chenhao Wang \thanks{This research was funded by the National Key R\&D Program of China under Grant number 2021YFB3900504.}
} 


\markboth{Journal of \LaTeX\ Template,~Vol.~14, No.~16, October~2023}%
{Shell \MakeLowercase{\textit{et al.}}: A Sample Article Using IEEEtran.cls for IEEE Journals}


\maketitle
\begin{abstract}
More accurate extraction of invisible building footprints from very-high-resolution (VHR) aerial images relies on roof segmentation and roof-to-footprint offset extraction. 
Existing methods based on instance segmentation suffer from poor generalization when extended to large-scale data production
and fail to achieve low-cost human interaction. 
This prompt paradigm inspires us to design a promptable framework for roof and offset extraction, and transforms end-to-end algorithms into promptable methods. 
Within this framework, we propose a novel Offset-Building Model (OBM).
Based on prompt prediction, we first discover a common pattern of predicting offsets and tailored Distance-NMS (DNMS) algorithms for offset optimization. 
To rigorously evaluate the algorithm's capabilities, we introduce a prompt-based evaluation method, 
where our model reduces offset errors by 16.6\% and improves roof Intersection over Union (IoU) by 10.8\% compared to other models. 
Leveraging the common patterns in predicting offsets, DNMS algorithms enable models to further reduce offset vector loss by 6.5\%. 
To further validate the generalization of models, we tested them using a newly proposed test set, Huizhou test set, with over 7,000 manually annotated instance samples. 
Our algorithms and dataset will be available at \url{https://github.com/likaiucas/OBM}. 
\end{abstract}

\begin{IEEEkeywords}
    Building footprint extraction, Roof segmentation, Roof-to-footprint offset extraction, Segment Anything Model (SAM), Non-Maximum Suppression(NMS)
\end{IEEEkeywords}
    
\section{Introduction}
\label{sec:Intro}
The problem of Building Footprint Extraction (BFE) has a history of over a decade, which can benefit 3D building modeling, 
building change detection, and building height estimation \cite{a1,a2,a3,a4,a5}, and there were many meaningful researches about cities\cite{landprice, urbanregion,huang2022evaluation}. 
Early studies extract building footprints by measuring geometric features and traditional machine learning methods\cite{a6}. 
These kinds of methods\cite{a7,a8} usually limited by discriminating shallow features and cannot be applied in a larger scale. 
Recently, deep learning based building related methods populate in solving BFE problems \cite{a9, a10, a11}.
This kind of methods focus on near-nadir images where building roofs and footprints are nearly vertical or roof-to-footprint offsets are very small. 
However, demanding shooting angles of satellites make near-nadir images hard to be derived. 
In other words, if models can extract footprints in off-nadir images, the production cost of obtaining remote sensing images through photography will be lowered. 

To solve this problem, Wang \etal\cite{a12} propose a model called Learning Offset Vector (LOFT) and a related dataset BONAI in 2023. 
LOFT provides a new idea to solve the BFE problem. 
It simulates footprint-labelling process of human, extracting a roof and a roof-to-footprint offset, and then utilizing both outputs, 
directly drag the roof to its footprint in the direction of offset. 
The LOFT model adopts a two-stage instance segmentation structure similar to Mask RCNN\cite{a13}, 
featuring an offset head based on convolution and Region of Interest Align (ROI Align). 
Under the similar idea of LOFT, Weijia Li proposed MLS-BRN\cite{MLSBRN}, 
a multi-level supervised building reconstruction network which allows more kinds of annotated building dataset involved in training the same model. 

\begin{figure*}[htbp]
	\centering
	\begin{minipage}{0.24\linewidth}
		\vspace{4pt}
		\centerline{\includegraphics[width=\textwidth]{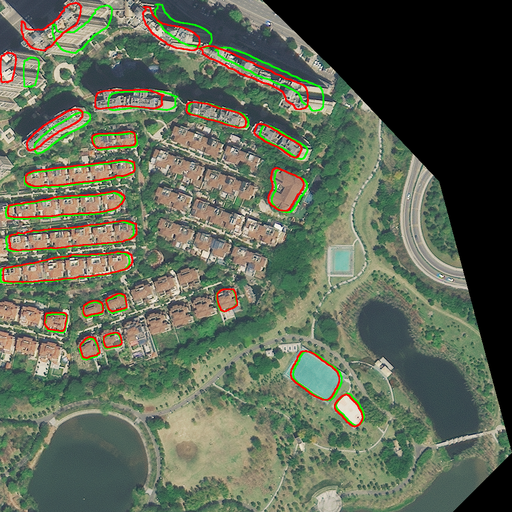}}
        \centerline{LOFT + NMS} 
	\end{minipage}
	\begin{minipage}{0.24\linewidth}
		\vspace{4pt}
		\centerline{\includegraphics[width=\textwidth]{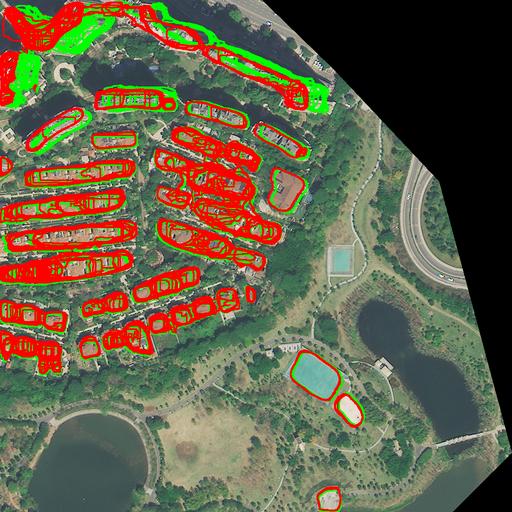}}
		\centerline{LOFT + soft NMS}
	\end{minipage}
	\begin{minipage}{0.24\linewidth}
		\vspace{4pt}
		\centerline{\includegraphics[width=\textwidth]{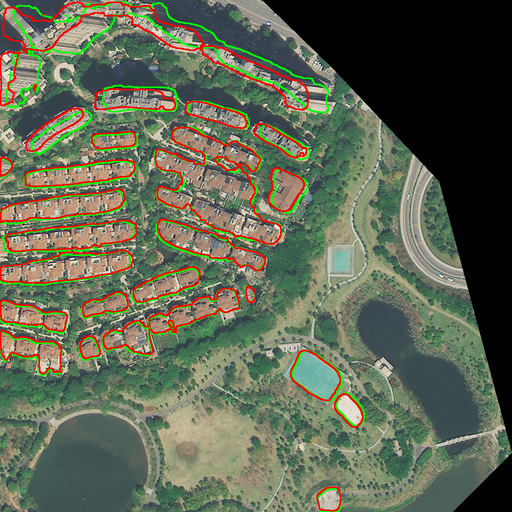}}
		\centerline{LOFT + soft NMS + Merge}
	\end{minipage}
    \begin{minipage}{0.24\linewidth}
		\vspace{4pt}
		\centerline{\includegraphics[width=\textwidth]{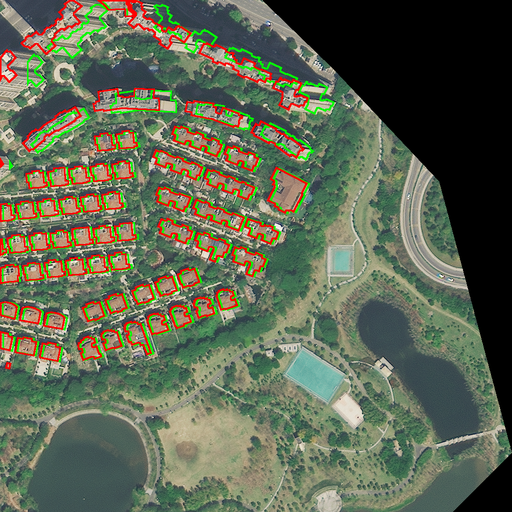}}
		\centerline{Ground Truth}
	\end{minipage}
	\caption{In given pictures, red boundaries and green boundaries represent roof and footprint respectively. During large-scale data production, instance segmentation methods face challenges related to generalization. 
	The listed results are from LOFT\cite{a12}, and the input images are from real production process which is {\bf 100\% unseen by LOFT}.
	Apart from mistake recognition, problems manifest in two aspects. 
	Firstly, these methods usually rely on post-processing algorithms. 
	Showing in first picture, a strict NMS algorithm lost many instances. 
	To address this, soft NMS\cite{a14} often applied to minimize the number of missing samples. 
	However, lower score thresholds of soft NMS consequently matched one building with many instances in pieces. 
	The confusing results let data producer hard to choose correct instances. 
	Of course, predicted buildings in neighbor can be merged and fused together as shown in the third picture. 
	However, the results let densely populated buildings stick together, far from getting wanted results as listed ground truth. 
	Secondly, data producers have to plot out those missing samples point-by-point, because of inflexible Region Proposal Network (RPN).}
	\label{fig.0}
\end{figure*}

However, when extending such method to large-scale data production, such idea suffers from poor generalization. 
Additionally, the unstable performance of NMS algorithms will make the outputs of these end-to-end models hardly be applied in data production.  
As illustrated in \cref{fig.0}, this image is from real data production, 
whose shooting location and other camera related information are unknown towards the model.
Within the given images, apart from plenty of mistakes made by instance-level models, 
hyperparameters of NMS often let data producers in a dilemma: higher output score thresholds of NMS will miss more samples, 
while lower scores let them hard to select correct instances. 

With the turn up of prompt paradigm, a cutting-edge model, namely Segment Anything Model (SAM)\cite{a16}, 
allows for powerful zero-shot capabilities with flexible prompting. 
SAM has applied widely in dataset labeling. Data producers only plot a point, a box or a coarse-labeled mask, then SAM can give a better mask label. 
Inherit from which, many works have proved its success. 
Fine-tuned from SAM, SAM-HQ improved segmentation quality of SAM by solving the problem of 16$\times$-downsampling feature maps\cite{a17}. 
RS-Prompter\cite{a18} can complete many downstream mask-related tasks by providing prompts to SAM. 
If we can control the input of prompts and the model will only give results in interested areas, the dilemma caused by end-to-end methods will be solved. 
Integrating how Wang\cite{a12} finds building footprints and powerful SAM, we are lighted to solve BFE problem in prompt level: \textit{“can we provide a prompt and a model directly plots the roof and drag it to its footprint like a human annotator?”}. 

Besides, NMS, soft NMS\cite{a14} and softer NMS\cite{a15} are designed for object detection and instance segmentation. They adjust intersecting boxes by boxes' score. However, simply deleting or adjusting box score cannot improve offset qualities. But the idea of soft NMS really inspired us, \textit{“can we proofread low-quality offsets based on those well-predicted offsets?”}.

\begin{figure}
	\centering
	\includegraphics[width=0.9\linewidth]{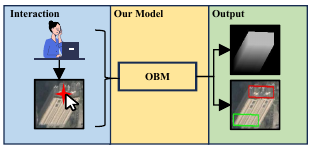}
	\caption{With provided prompts, our model can extract roof and footprint for buildings and generate a relative height map.}
	\label{fig.00}
\end{figure}

To answer mentioned two questions, we devised a workflow inspired by SAM, transforming Wang's models\cite{a12} into a promptable model compatible with box prompts. 
Additionally, leveraging SAM, we proposed Offset Building Model (OBM)\cref{fig.00}. 
To tackle problems of NMS algorithms, we identified general patterns by experiments, leading to the creation of the DNMS algorithms. 
DNMS will not delete instances, but improve offsets quality in reality. 
We also established a novel evaluation system for prompt-level models, by which we can depict ability of models in more details. 
By conducting extra experiments on a new dataset, we proved the superiority of our OBM and DNMS algorithms. 
Moreover, with accurate extraction of roofs and footprints, another contribution of our work can influence the process of monocular height estimation. 
Height information can be retrieved directly with 3D-aware techniques, e.g., 3D sensors such as light detection and ranging (LiDAR)\cite{a19, a20, a21}. 
However, rare data derived by such conditionally applicable sensors cannot fulfill the demand of related deep learning methods\cite{a5}, but
our methods can generate relative height map of buildings using optical remote sensing images. 
Different from above-mentioned height maps, the only gap between height maps and our maps can be bridged by a scale factor or camera parameters. 

In summary, our contributions are as follows:
\begin{enumerate}
 \item[$\bullet$] We proposed a prompt-level footprint model OBM, and OBM was the first model which introduced the offset tokens and ROAM structure to predict offsets. 
 \item[$\bullet$] Based on the idea of prompt, we transformed open-source RoI-based methods into prompt mode. 
 \item[$\bullet$] We designed DNMS algorithms to improve the quality of predicted offsets. The algorithms are based on the common pattern of predicting offsets and can effectively improve the directional quality of offsets. 
 \item[$\bullet$] We presented a comprehensible and detailed metric method for prompt-level offset and roof regression. 
\end{enumerate}

\section{Related Work}
\label{sec:SAMapplications}
OBM can be seen as an application of a large vision model (SAM) and multitask learning to solve BFE problem. 
Similar with object query,  OBM uses an offset query to represent roof-to-footprint offset. 

{\bf Segment Anything Model and Applications:} 
Segmentation is a fundamental task in computer vision\cite{b31,b32,b33,b34,b35,b36,b37,a16}, which requires a pixel-wise understanding of a image.
SAM is a powerful zero-shot segmentation model\cite{a16}, by using minimal human input, such as bounding boxes, reference points, or simply text-based prompts.
Based on SAM, Fast-SAM\cite{b34} realized a 50x speedup by using a CNN-based decoder. 
PerSAM can segment personal objects in any context with favorable performance, and PerSAM-F further alleviates the ambiguity issue by scale-aware fine-tuning\cite{b29}. 
PerSAM can also improve DreamBooth\cite{b30} by mitigating the disturbance of backgrounds during training. 
SAM-PT\cite{b38} prompts SAM with sparse point trajectories predicted using point trackers, such as CoTracker\cite{b39}. 
SAM-HQ\cite{a17} is a high-quality version of SAM, which can generate high-resolution masks. This improvement is achieved by introducing negligible overhead to the original SAM. 
In SAM-HQ, a lightweight High-quality Output Token in HQ-SAM to replace the original SAM’s output token for high-quality mask prediction. 
SAM-Adapter\cite{b40} proposes the adaptation approach to adapt SAM to downstream tasks and achieve enhanced performance.
After SAM launched, it has been used in remote sensing \cite{b28,a18}. 
In \cite{b28},  involves using a prompt-text-based segmentation as a training sample (instead of a human-labeled sample), making it an automated process for refining SAM on remote sensing imagery. 
RSprompter\cite{a18} proposes a novel prompt learning method that augments the SAM model's capabilities, thereby facilitating instance segmentation in remote sensing imagery.
SAM-RBox can detect a rotated box for different items in remote sensing images, which uses a trained horizontal Fully Convolutional One-Stage object (FCOS) detector to provide HBoxes into SAM as prompts \cite{yu2023h2rboxv2}. 
SAM has not be applied in building footprint extraction, and OBM is the first work to apply SAM in BFE problem.

{\bf BFE Problem Solutions:} 
Spatial residual inception (SRI) module\cite{b41} was proposed to capture and aggregate multiscale contextual information, and this module could improve the discriminative ability of the model and obtain more accurate building boundaries. 
Jian Kang\cite{b42} investigates the problem of deep learning-based building footprint segmentation with missing annotations, approaching it from the perspective of designing an effective loss function to specifically deal with this issue.
Weijia Li\cite{b43} explored the combination of multi-source GIS map datasets and multi-spectral satellite images for building footprint extraction in four cities using a U-Net-based semantic segmentation model for building footprint extraction. 
ConvBNet\cite{b44} and MHA-Net\cite{b45} was tailored to the complex textures, varying scales and shapes, and other confusing artificial objects in building footprint extraction. 
Apart from that, building segmentation is also a related task\cite{darnet,ma2023local,shi2020building,sariturk2022residual}. DARNet\cite{darnet} using proposed loss function to directly encourages the contours to match building boundaries. 
Zhanming Ma \emph{et al.}\cite{ma2023local} proposed a local feature search network with discarding attention module (DFSNet) to help the model distinguish the building areas and water bodies. 
Yilei Shi \emph{et al.}\cite{shi2020building} integrating graph convolutional network (GCN) and deep structured feature embedding (DSFE) into an end-to-end workflow to overcome the issue of delineation of boundaries. 
RIU-Net \cite{sariturk2022residual}, a residual connected, Inception-based, u-shaped encoder-decoder architecture with skip connections, was proposed to segment buildings in remote sensing imagery. 
In offset-based methods, Gordon Christie\cite{christie2020unet} introduced a U-Net based methods to predict image-level orientation as \(sin(\theta)\) and \(cos(\theta)\). 
With this map, a given ground truth footprint or roof can finally be dragged to its related roof or footprint.
Using a similar idea, MTBR-Net\cite{MTBRNet} using semantic related tasks and offset related tasks to extract 3D buildings and was equipped with more functions. 
Based on a similar idea of offsets, LOFT\cite{a12} uses instance-level offset as a feature to describe buildings. 
In CVPR 2024, MLS-BRN\cite{MLSBRN} adds new tasks to bridge the gaps between different building instance, which finally alleviates the demand on 3D annotations. 
Different with aforementioned, OBM supports a prompt-level footprint extraction and using DNMS algorithms to improve the prediction quality of offsets. 

{\bf Object Query:}
Queries, stored as learnable vectors, are widely used to imply properties of instance\cite{a27, a23,maskformer,masktwoformer,dn, dino, maskdino}. 
Object query for detection was first proposed in DETR\cite{a23}, which is used to represent the bounding box area. 
In DETR, the model will pre-set $n\times$ object queries for $n$ instances, and, at the final stage, these $n$ queries will be decoded as $n$ bounding boxes with several layers of  Feed-Forward Network (FFN).
Followed by that, MaskFormer\cite{maskformer} and Mask2Former\cite{masktwoformer} are another series of transformer model for semantic segmentation, using queries to represent instances. 
Both models use a single head to predict binary masks and a classification head to predict the class of each mask layer.
Dynamic Anchor Boxes DETR(DAB-DETR)\cite{a27} offers a deeper understanding of the role of queries in DETR. DeNoising DETR(DN-DETR)\cite{dn} offered a deepened understanding of the slow convergence issue of DETR-like methods, 
and using a novel denoising method related with queries to speedup training. 
Based on aforementioned ideas, the same team proposed DINO\cite{dino} for object detection and MaskDINO\cite{maskdino} for instance segmentation, semantic segmentation and panoptic segmentation. 
OBM is the first work to use offset query and Reference Offset Augment Module (ROAM) structure to predict roof-to-footprint offset. 

{\bf Multi-Task Learning and Deep Learning in Remote Sensing:}
Multi-Task Learning (MTL) is a learning paradigm in machine learning and its aim is to leverage useful information
contained in multiple related tasks to help improve the generalization performance of all the tasks \cite{zhang2021survey}. 
Multitask model can do more things using the same model \cite{htc, a12, li2020automatic, eem, drive, liu2021abnet}. Hybrid Task Cascade (HTC)\cite{htc} effectively integrates cascade into instance segmentation by interweaving detection and segmentation features together for a joint multi-stage processing.
LOFT\cite{a12} can predict roof-to-footprint offset with roof segmentation. 
P. Li \emph{et al.}\cite{li2020automatic} proposed a method which can extract road and road intersections. 
Eagle-Eyed Multitask CNNs\cite{eem} incorporate three tasks, aerial scene classification (ASC), center-metric learning and similarity distribution learning together. 
For autonomous driving, a shared backbone with different subnets was designed to address object detection, drivable area segmentation, lane detection, depth estimation and so on \cite{drive}.
HybridNet \cite{hybridnet} is one of those models developed an end-to-end perception network to perform multi-tasking, including traffic object detection, drivable area segmentation and lane detection. 
Additionally, YOLOP \cite{yolop} and YOLOPv2 \cite{yolopv2} further enhanced the performance of aforementioned three tasks by using the share-weight backbone. 
ABNet\cite{liu2021abnet} was designed to solve challengings caused by complicated backgrounds and imbalanced scale and distribution of remote sensing image. 
Salient object detection in optical remote sensing images often suffers from intrinsic problems, such as cluttered background, scale variation and irregular topology, to solve this HFANet\cite{wang2022hybrid} was proposed. 
SDNet\cite{SDNet} is a dual-branch network which can perform cross-task knowledge distillation from the scene classification to facilitate accurate saliency detection. 
\section{Methodology}
\cref{subsec:1} introduces the problem formulation.
\cref{subsec:2} describes the OBM model.
\cref{subsec:3} introduces DNMS and soft DNMS algorithms.
\cref{subsec:4} describes ROI prompt based models.
\cref{subsec:5} describes how to generate relative height map with roof segmentation and offsets.
\cref{subsec:6} introduces metric methods.
\subsection{Problem formulation}
\label{subsec:1}
Instance segmentation methods only need an image as input, and the model will output roofs and offsets. 
In contrast, prompt-level models require an input image along with a series of prompts. 
The model will then output roofs and offsets of the building within the area indicated by prompts. 
\begin{equation}
  \begin{cases}
  D = \{(I_i, P_i, T_i); i=1,2,\ldots,N\} \\
  P_i \in \{\text{bbox, point, mask}\} \\
  T_i = MODEL(I_i, P_i), \quad i=1,2,\ldots,N \\
  \end{cases}
  \label{eq:1}
\end{equation}
In the above equation, \( D \) represents the dataset BONAI\cite{a12}. 
\( I_i \) denotes the $i$ th image, \( P_i \) represents the prompt for the $i$ th image, 
and \( T_i \) represents the regression target, including a roof segmentation and offset, which were related to prompt \( P_i \) for the $i$ th image, 
typically including the roof mask and offsets. The $MODEL$ represents functions that used to solve BFE problem. 
In input stage, prompt masks and points are sampled from the ground labels' masks and points.
\subsection{Offset Building Model (OBM)}
\label{subsec:2}
OBM can predict roof-to-footprint offsets, roof segmentation, and building segmentation for each inputted image and its bounding box prompts. 
Using predicted offsets and roofs, a relative height map and building footprints can be figured out.

SAM\cite{a16} is composed of three parts: 
(a) Image Encoder: an MAE\cite{a22} pretrained ViT-based backbone which is used to extract image features; 
(b) Prompt Encoder: encode interactive boxes/points/masks into different tokens to imply areas of input image; 
(c) Mask Decoder: a two-layer two-way transformer, 
which combines image features and tokenized prompts, and finally give out final masks. 

\begin{figure*}[htbp]
  \centering
    \includegraphics[width=1\linewidth]{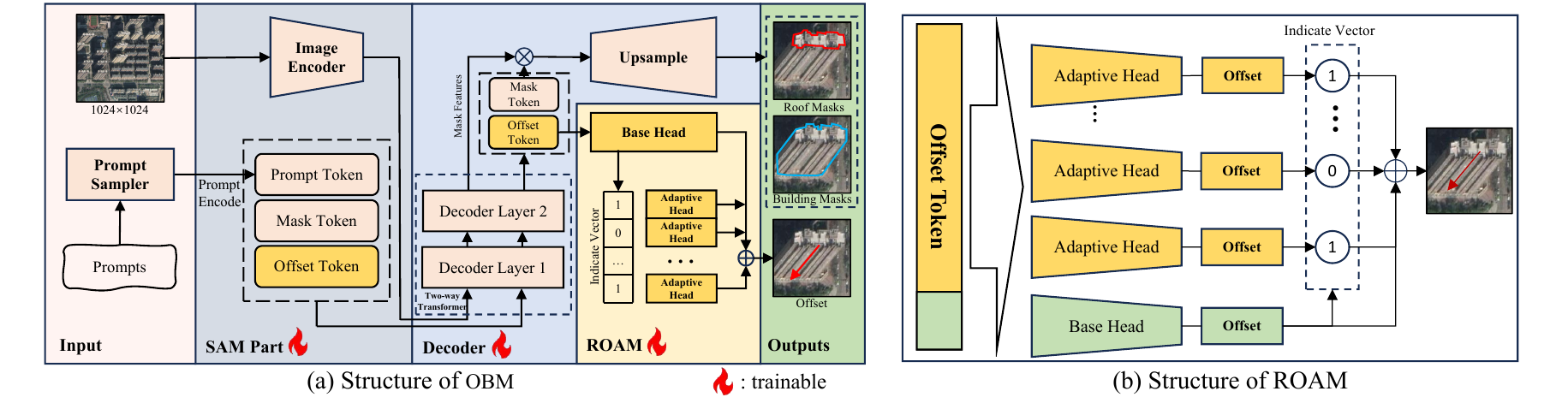}
  \caption{In (a), the OBM extends the SAM model by adding an offset prediction branch, namely ROAM. To adapt to diverse GPU capacities, 
  we implemented an optional Prompt Sampler for prompt selection. The Offset Tokens, along with Prompt Tokens and Mask Tokens are fed into the Decoder. Additionally, an Offset Coder similar to DETR's Box Coder enhances offset training. 
  As shown in (b), ROAM are used for offset prediction, which are composed by a Base Head and several Adaptive Head. 
  Base Head will firstly generate a reference offset and an indicator vector.
  Indicator vectors from the Base Head select offsets and then roam offsets to different Adaptive Offset Heads. 
  The ultimate output offsets are derived from both base head and adaptive heads. }
  \label{fig:fig}
\end{figure*}

As shown in \cref{fig:fig}, OBM inherits from SAM's segmentation ability, and we implement a novel Reference Offset Augment Module (ROAM) to predict roof-to-footprint offsets. 
ROAM includes a series of offset tokens, FFN and offset coders. 
To match different capability of GPUs and accelerate training, we designed a Prompt Sampler. 
In OBM, we preserve most of SAM structures for its ability of perceiving objects. 
Since the roof and building have sharing semantic pixels, what we need to do is ensuring that each mask token focuses solely on the roof or the building. 
In other words, eliminating cognitive ambiguity in models. 

\subsubsection{Prompt Sampler}
Similar with object detection and instance segmentation, our roof segmentation and offset prediction also have their “tiny object problem\cite{a24}”. Via huge number of experiments, we discover that low buildings will slow down the convergence of our model. To solve this problem, we design this prompt sampler to mitigate the influence of low buildings. Apart from this, considering limited computing resources, we design this part to cut the demanding usage of GPUs. Currently, the paper provides two sampling modes:

{\bf Uniform Random Sampling:} In this mode, prompts are randomly sampled from the entire set with same probability. 
This approach ensures a diverse selection of prompts for training.

{\bf Length-based Sampling:} In this mode, prompts are sampled with probabilities based on lengths of roof-to-footprint offset. 
Offset length influences the likelihood of selection, allowing prompts with different offset lengths to be included in training. 
This method ensures a more balanced representation of prompts.
\subsubsection{Offset Token and Offset Coder}
Offset Token and Offset Coder are core of ROAM. Similar to the object query in DETR\cite{a23}, we created offset tokens for offset regression. 
This offset token, derived from the decoder output, is processed through an FFN and then utilized by the offset coder to obtain the final offset output.

We employ the following encoding approach for roof-to-footprint offsets during training and inferring:
\begin{equation}
  \begin{cases}
   \mathcal{O}_e = \frac{\mathcal{O} - \alpha\times \beta}{\alpha \times \gamma}\ \\
   \mathcal{O} = (\gamma \times \mathcal{O}_e + \beta) \times \alpha \ \\
\end{cases}
\label{eq:2}
\end{equation}
In \cref{eq:2}, \( \alpha \) represents the offset scaling factor, 
while \( \beta \) and \( \gamma \) are parameters used for normalizing the results. 
During the training phase, the model regresses the \( \mathcal{O}_e \) values. 
This system is employed to calculate the final offset \( \mathcal{O} \). 
During the inference phase, final offsets are obtained through the decoding process.
\subsubsection{Reference Offset Augment Module (ROAM)}
In DETR\cite{a23}, the box coder uses the size of input images to scale down the output results, enabling better fitting for the FFN. 
However, this strategy cannot be applied to the Offset Coder. 
Based on our training experience, building offsets in the $x$ and $y$ directions often approach zero. 
If the Offset Coder copied encoding and decoding processes of DETR's Box Coder entirely, the loss would become even closer to zero. 
We can deepen our understanding of this phenomenon through formulas in \cref{eq:2}.
The \( \alpha \) is a constant in a box coder, and can be understood as a scale factor.
According to statistics, most buildings have an offset length of less than 40 pixels.
If \( \alpha \) is too large, the most of \(\mathcal{O}_e\) used in training will fluctuate around 0, and that means the model will learn nothing. 
Similarly, if \( \alpha \) is too small, the value range of \(\mathcal{O}_e\) will be extremely unstable, which will lead to unstable convergence of the model.
However, larger \( \alpha \) and smaller \( \alpha \) are valuable for the learning of longer offsets and shorter offsets. 
To improve the offset quality, we proposed ROAM and combined aforementioned coders in adaptive heads. Each head in ROAM is similar except the scale factor of Offset Coder. 

The workflow of the ROAM module, as depicted in \cref{fig:fig}, 
for each prompt, base offset head will figure out a rough offset length, 
and generate an indicator for it to choose suitable adaptive heads aiding final results. 
Mathematically, this can be expressed as:
\begin{equation}
    \mathcal{O}^i = \frac{\mathcal{O}_R^i + \sum_{i=1}^{n} \omega^i \mathcal{O}_A^i}{1 + \sum_{i=1}^{n} \omega^i}
\label{eq:3}
\end{equation}
Here, \( \mathcal{O}_R^i \) represents the \(i\) th Reference Offset, 
\( \mathcal{O}_A^i\) represents the \(i\) th Adaptive Offset, 
and \( \omega^i \)   is the indicator value, taking values of 0 or 1, 
indicating whether the corresponding Adaptive Offset affects the output $\mathcal{O}^i$. 
The average of the Reference Offset and $n$ Adaptive Offsets is the final output.
The whole computing process can be described as: at the inference stage, Base Head, whose $\alpha$ in Offset Coder is moderate, will first give out a rough offset for each building. 
These offsets can be provided as references for Adaptive Heads, and, based on their lengths, they can be roughly divided into two groups: longer offsets and shorter offsets. 
Each group will use a different indicator to identify the offsets.  
Then, both groups will be assigned to different Adaptive Heads, and their assignments commonly follow the rule: 
longer offsets will use Adaptive Head with larger $\alpha$ and heads with smaller $\alpha$ for shorter offsets. 
With multi-heads outputs, the offsets will be finally determined.
In \cref{sec:roam}, we will introduce the detailed settings in our released model version. 

The OBM model is finally obtained via minimizing a joint loss function,
\begin{equation}
\mathcal{L} = \mathcal{L}_{ROAM} + \mathcal{L}_{roof} + \mathcal{L}_{building} 
\label{eq:4}
\end{equation}
where \( \mathcal{L}_{ROAM} \) is the loss of ROAM, \( \mathcal{L}_{roof} \) is CrossEntropy Loss \cite{celoss} of roof segmentation, and \( \mathcal{L}_{building} \) is CrossEntropy Loss of building segmentation.

\begin{equation}
  \mathcal{L}_{ROAM} = \mathcal{L}_B + \sum_{i=0}^{n} \mathcal{L}_i 
\end{equation}
\(\mathcal{L}_i\) represents offset loss of the \(i\) th adaptive offset head. \(\mathcal{L}_B\) represents offset loss from reference offset head. SmoothL1 Loss \cite{fastrcnn} is applied for all offset heads.
\subsection{Distance NMS and soft Distance NMS}
\label{subsec:3}

Distance NMS (D-NMS) and soft Distance NMS (soft D-NMS) algorithms are specifically designed for refining building offset predictions. 
As shown in \cref{tab.1}, big buildings often give out a more accurate offset in direction compared with shorter. Therefore, drawing on the concepts of NMS and soft NMS algorithms, we propose an NMS algorithm based on the predicted offset lengths.

D-NMS directly replaces directions of all offsets by the longest. But soft D-NMS is different, because it will use statistics to adjust
the longer and shorter offsets. 

\begin{algorithm}[!hbp]
	\caption{soft D-NMS}
    \label{sdms}
    \begin{algorithmic}
      \REQUIRE Offsets: $O:\{\theta_i, \rho_i, \vec p_i=(x_i, y_i) | i = 1,2,...,n\}$
        \STATE The average length of $O$: $\mu \gets \frac{\sum_{i=1}^{n} \rho_i}{n}$ 
        \STATE The standard of $O$: $\sigma^2 \gets \mu \frac{\sum_{i=1}^{n} (\rho_i - \mu)^2}{n-1}$  
        \STATE Reference length:  $r = \mu+ \alpha \times \sigma$
        \STATE Gaussian Distance: $\mathcal{D}:\{d_i = e^{-\frac{(\rho_i - r)^2}{2\sigma^2} }| i = 1,2,...,n\}$
        \STATE Standard direction: $x_s = \frac{\sum_{i=1}^{n} \omega_i x_i}{\sum_{i=1}^{n}\omega_i}$; $y_s = \frac{\sum_{i=1}^{n} \omega_i y_i}{\sum_{i=1}^{n}\omega_i}$
        \STATE Unit direction: $\vec p_u = (x_u, y_u)$, \\
              \hspace{2em}where: $x_u = \frac{x_s}{\sqrt{{x_s}^2 + {y_s}^2}}$; $y_u = \frac{y_s}{\sqrt{{x_s}^2 + {y_s}^2}}$
        \STATE Fixed Offset: $\vec f_i = (1-d_i)\times \vec p_i + d_i \rho_i \times \vec p_u$
        \ENSURE Offsets: $O:\{\vec f_i | i = 1,2,...,n\}$
    \end{algorithmic}
\end{algorithm}
In \cref{sdms}, the offset values expressed in both Cartesian and polar coordinates will be used as inputs, and the algorithm will output the final corrected offset result. 
The \( \alpha \) is a constant whose value is around 0.1. 
The \( \omega_i \) equal to 1 when the length of \( \rho_i \) ranked in top-k otherwise 0.

\subsection{ROI Prompt Based Offset Extraction}
\label{subsec:4}
\begin{figure}[H]
  \centering
  \includegraphics[width=\linewidth]{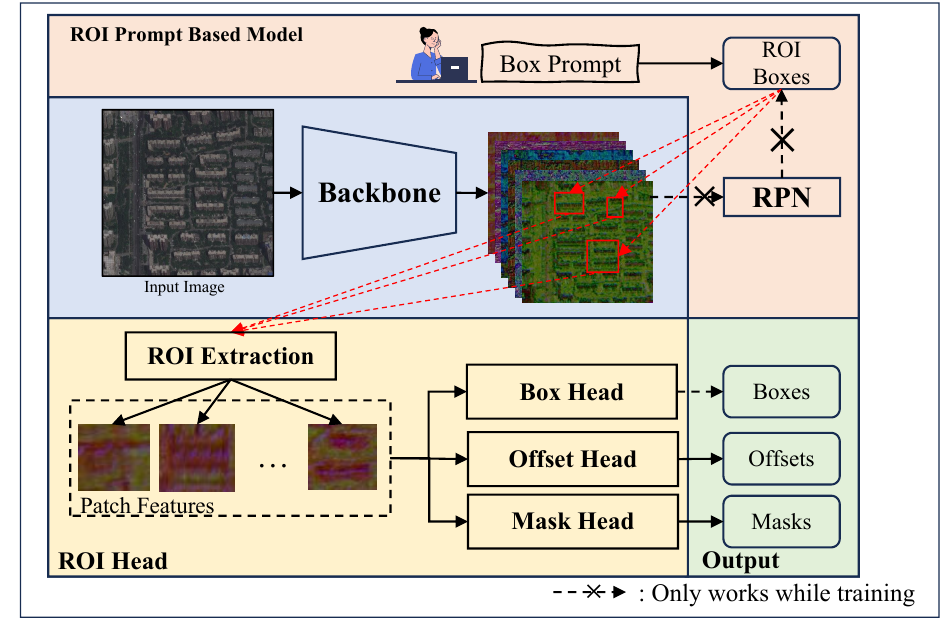}
  \caption{During the training of models, 
  RPN (Region Proposal Network) is utilized to produce boxes. 
  These boxes are employed with ROI Align to crop features, which is subsequently transformed into local features of identical size. 
  These features are then inputted into multitask heads for regression.
  In the inference phase, the RPN is deactivated, and models use manually provided boxes for ROI extraction.}
  \label{fig:3}
\end{figure}

In two-stage models, inflexible RPN gets in the way of human interaction. In \cref{fig:3}, following the idea of SAM\cite{a16}, we shield RPN at the inference stage to receive prompt boxes.
In the following part, we use LOFT to represent prompt LOFT, and cascade LOFT to represent prompt cascade LOFT.

To ensure the best training strategy, ROI prompt based models are trained using both prompt and RPN. 

\subsection{Relative Height Map}
\label{subsec:5}
\begin{figure}[htbp]
	\includegraphics[width=\linewidth]{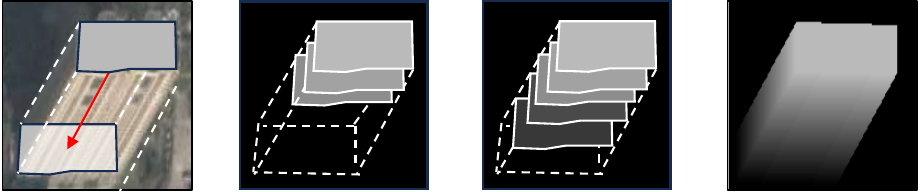}
	\caption{
    Relative height map will be generated by fading the top roof to its footprint.
  }
  \label{fig.rhm}
\end{figure}

With accurate roof segmentation and offset prediction, generating a relative height map is reliable. 
In the same image, all buildings will be allocated a relative height based on the length of their offsets.
As shown in \cref{fig.rhm}, the longest one will be defined as 1, at the inference stage, the top roof will be dragged to its footprint, 
and linearly interpolate the relative height.

The relative height of each building is determined by the length of their offsets. 
\begin{equation}
  h_i = \frac{\rho_i}{\max(\rho)}
\label{eq:height}
\end{equation}
where \( h_i \) represents the relative height of the \( i \) th building,
\( \rho_i \) represents the length of the \( i \) th building's offset, and
\( \max(\rho) \) represents the length of the longest offset in the given area.

As the relationship between each building can be defined by \cref{eq:height}, 
the height map can be generated by finding a scale factor. 
This factor can be determined by directly measuring a height of a building in the given area, 
or be figured out by parameters of shooting camera and satellite.

\subsection{Metric method}
\label{subsec:6}
\subsubsection{Offset metrics}

\begin{figure}[H]
  \centering
   \includegraphics[width=0.8\linewidth]{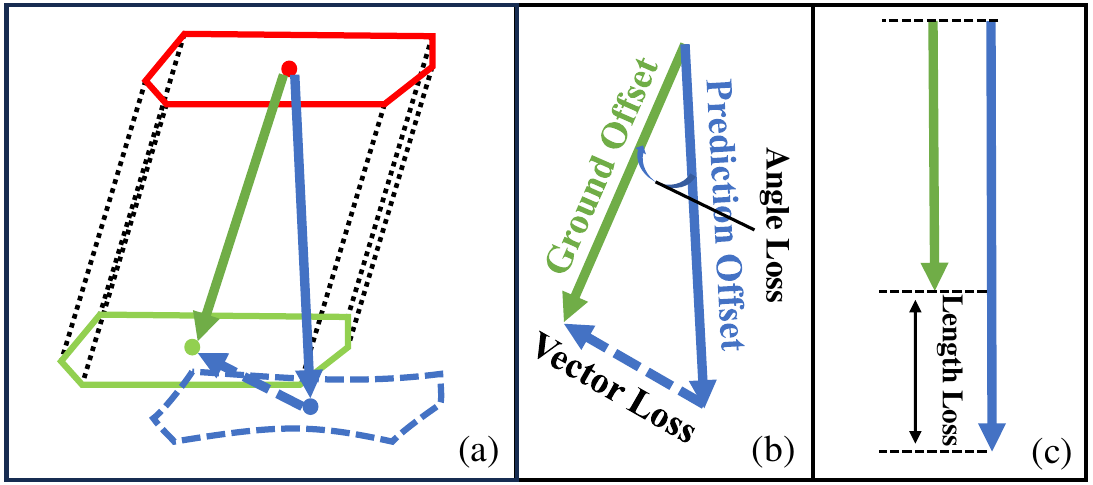}
   \caption{Offset Evaluation Illustrations: (a) demonstrates how we locate the footprints of the roof through the roof to footprint segmentation during offset prediction. (b) and (c) display three important metrics proposed by us to evaluate the losses in offset prediction.}
   \label{fig:4}
\end{figure}

As shown in \cref{fig:4}, we will measure each offset in three aspects, Vector Loss ($VL$), Length Loss ($LL$) and Angle Loss ($AL$). 
\begin{equation}
  VL = \left|\vec p - \vec p_g \right|_2;\\
  LL = \big| \left|\vec p\right|_2 - \left|\vec p_g \right|_2\big|;\\
  AL = |\theta - \theta_g|\\
\end{equation}
where \( \vec p \) and \( \vec p_g \) represent the predicted and ground truth offset.
\( \theta \) and \( \theta_g \) represent the predicted and ground truth angle.
The \( \left|\cdot\right| \) and \( \left|\cdot\right|_2 \) represent the 1-Norm and 2-Norm respectively.
Referring COCO metric\cite{a25}, we measure offsets based on different length group, and calculate the average of them. Above three losses, can describe as,
\begin{equation}
   m\mathcal{L} = \frac{\mathcal{L}_{(10n,\infty)} + \sum_{i=0}^{n} \mathcal{L}_{(10i,10i+10)}}{n+1} \\
\end{equation}
\noindent where \(\mathcal{L}\) can represent $VL$, $LL$ and $AL$.
\(\mathcal{L}_{(i,j)} \) represents average loss of offset whose ground truth length is between $i$ and $j$ pixels. 
We also use $aL$ to denote \(\mathcal{L}_{(0,\infty)}\). The units of $VL$ and $LL$ are pixels, while the units of $AL$ are angles in radians.

\subsubsection{Mask metrics}
Similar to evaluating offset errors at the prompt level, we compute IoU and Boundary IoU (BIoU) \cite{a26} for each roof mask with ground truth, 
and then calculate the mean values of all. In experiments, we use IoU and BIoU to represent each. 
For direct measurement of footprint quality, prompt-level F1score is also calculated.

\section{Experiment}
\begin{table*}[t]
    \centering
  \caption{Main results of each box prompt models tested on BONAI\cite{a12}.}
  \label{tab.1}
  \resizebox{\linewidth}{!}{
  \begin{tabular}{cc|ccccccccccccccc}
    \toprule
    {\textbf{Model}} & ~ & \textbf{0,10} & \textbf{10,20} & \textbf{20,30} & \textbf{30,40} & \textbf{40,50} & \textbf{50,60} & \textbf{60,70} & \textbf{70,80} & \textbf{80,90} & \textbf{90,100} & \textbf{100,\(\infty\) } & \textbf{$m$\(\mathcal{L}\)} & \textbf{$a$\(\mathcal{L}\)} &\textbf{IoU} &\textbf{BIoU}\\ \hline
        ~& $VL$ & 5.48 & 4.34 & 6.39 & 6.80 & 8.41 & 11.8 & 15.6 & 20.3 & 17.2 & \textbf{18.2} & \textbf{53.9} & 15.4 & 6.12 &~&~\\ 
        LOFT & $LL$ & 4.15 & 2.75 & 4.51 & 5.18 & 6.52 & 8.70 & 12.2 & 16.7 & 12.9 & \textbf{14.4} & \textbf{49.9} & 12.6 & 4.51& 0.65 & 0.38\\ 
        ~& $AL$ & 0.55 & 0.21 & 0.17 & 0.11 & 0.10 & 0.14 & 0.15 & 0.16 & 0.13 & 0.13 & 0.15 & 0.18 & 0.32 &~&~\\ \hline
        \multirow{3}{*}{\thead{Cas.\\LOFT}}& $VL$ & 5.62 & 4.12 & \textbf{5.42} & 6.11 & 7.90 & 12.8 & 17.3 & 22.7 & 17.9 & 18.7 & 55.3 & 15.8 & 5.97&~&~   \\ 
         &$LL$ & 4.28 & 2.66 & \textbf{3.71} & 4.72 & 6.30 & 10.8 & 15.2 & 20.2 & 14.1 & 14.8 & 51.3 & 13.5 & 4.48 &0.68&0.40\\
         &$AL$ & 0.55 & 0.19 & 0.15 & 0.09 & 0.09 & 0.14 & 0.13 & 0.13 & 0.14 & 0.15 & 0.14 & 0.17 & 0.31 &~&~\\ \hline
         ~& $VL$ & \textbf{4.01} & \textbf{3.71} & 5.55 & \textbf{6.10} & \textbf{7.58} & \textbf{9.18} & \textbf{12.5} & \textbf{16.9} & \textbf{15.1} & 21.2 & 61.4 & \textbf{14.8} & \textbf{5.08} &~&~\\ 
        Ours& $LL$ & \textbf{3.20} & \textbf{2.55} & 4.14 & \textbf{4.96} & \textbf{6.14} & \textbf{7.77} & \textbf{10.8} & \textbf{15.9} & \textbf{13.8} & 19.5 & 60.1 & \textbf{13.5} & \textbf{4.03} &\textbf{0.73}&\textbf{0.43}\\
        & $AL$ & \textbf{0.37} & \textbf{0.15} & \textbf{0.14} & \textbf{0.08} & \textbf{0.08} & \textbf{0.07} & \textbf{0.08} & \textbf{0.06} & \textbf{0.06} & \textbf{0.10} & \textbf{0.11} & \textbf{0.12} & \textbf{0.22} &~&~\\ \bottomrule
    \end{tabular}
  }
\end{table*}
Although OBM supports multi-kinds of prompts but ROI model cannot. To make comparison, 
we focus on box prompt in experiment part. 
In first part, we conduct experiments on BONAI dataset\cite{a12} and test DNMS algorithms for different models.
Then we will test the generalization ability of models on a new dataset. All models are trained on MMDetection platform\cite{mmdetection}.

Our OBM was trained on a server with 6 RTX 3090 for over 12 hours.
On each GPU, we have 1 sample; in other words, the Batch Size during training was 6. 
We choose SGD \cite{sgd} as training optimizer, and the learning rate, momentum, and weight decay are 0.0025, 0.9 and 0.0001 respectively. 
GradClip \cite{gradclip} was used to ensure the model was trained on the right way at the beginning. 

The training process of OBM is divided into two stages.
First stage mainly concentrates on those long offsets, the next stage trained on all data. 
Both stage needs 48 epochs. 

\subsection{Main result}
The main results will focus on the models' performance on BONAI dataset\cite{a12}. This dataset includes 3,000 and 300 images (shape 1024$\times$1024) for train-val and test respectively. 

\cref{tab.1} illustrates big performance difference between each model, but all models are inclined to predict better direction for buildings with longer offset.
\{$i, j$\} in the first line represents the length range of ground offset.
In the first 11 columns, we list model performance in different length ranges.
The performance of our model exceeds all other models. For buildings with offset over 90 pixels, OBM predicts a little worse than LOFT ($VL_{(90,100)}$ increased by 3 pixels). 
This might be caused by the distribution of training data. On comprehensive level, OBM dropped $mVL$, $mAL$, $aVL$ and $aA$ by 3.8\%, 33.3\%, 17.0\% and 31.3\% compared with prompt LOFT. 
Moreover, because OBM's mask prediction ability was inherited from SAM, the roof performance of OBM significantly outperforms other models was within estimation. 
For example, roof IoU between predictions and Ground Truth (GT) improved by 12.3\% compared with that of prompt LOFT.

\cref{tab.22b} shows DNMS algorithms still work well by correcting offset angles. 
While OBM already has a good ability to distinguish offset angles, the improvement of DNMS is limited. 
DNMS algorithms can also improve performance on the new dataset. DNMS and soft DNMS model can correct \(aAL\) by 0.04-0.44 for different models. 
The improvements are related with the model's direction performance. 
\eg OBM has a very good directional perception, improvements made by correcting angles using DNMS and soft DNMS are limited ($mAL$ and $aAL$ dropped only around 0.01). 
While these improvements on prompt LOFT and prompt Cascade LOFT are very significant, \eg soft DNMS dropped $mAL$ and $aAL$ by 22.2\% and 28.1\% for prompt LOFT, and consequently $mVL$ and $aVL$ improved.

The improvements seem different from the huge gap between NMS and soft NMS, a supervising result can be found comparing \cref{tab.22b} and \cref{tab.22a}: 
soft DNMS algorithm cannot always defeat the performance of DNMS. To understand this, we need to know the difference of their mechanisms. 
Soft NMS allows models to output more bounding boxes with low confidence, 
and end-to-end models are inclined to predict only a few number of valuable bounding boxes with higher scores. 
In other words, soft NMS can help the models recall more correct bounding boxes. However, soft DNMS and DNMS are different. 
Both methods mainly operate the direction of predicted offsets to correct results. Obviously, this process will not delete any bounding boxes.
The final results will only be determined by the global performance and the performance of the tallest building respectively. 
Thus, there is a possibility that the longest offsets outperforms the global results in terms of a certain dataset at the most of the time. 
That is the reason why soft DNMS cannot always defeat DNMS. 

\begin{table}[htp]
  \centering
  \caption{Positive influence of DNMS algorithms on BOANI\cite{a12}.}
  \label{tab.22b}
  \resizebox{\linewidth}{!}{
  \begin{tabular}{c|rrrrrr}
  \toprule
  \textbf{Model} & \textbf{$mVL$} & \textbf{$mLL$} & \textbf{$mAL$} & \textbf{$aVL$} & \textbf{$aLL$} & \textbf{$aAL$} \\ \hline
      LOFT & 15.36&\textbf{12.59}&0.18&6.12&4.51&0.32\\ 
      +DNMS & \textbf{14.81} & 12.62 & 0.14 & \textbf{5.68} & 4.50 & 0.23 \\ 
      +s.DNMS &14.84&12.62&\textbf{0.14}&5.69&\textbf{4.50}&\textbf{0.23} \\ \hline
      Cas.LOFT & 15.80 & \textbf{13.45} & 0.17 & 5.97 & 4.48 & 0.31 \\ 
      +DNMS & 16.03 & \textbf{13.45} & 0.14 & 5.88 & 4.48 & 0.24 \\ 
      +s.DNMS & \textbf{15.24} & 13.48 & \textbf{0.12} & \textbf{5.58} & \textbf{4.47} & \textbf{0.23} \\ \hline
      Ours & 14.84 & 13.53 & 0.12 & 5.08 & 4.03 & 0.22 \\ 
      +DNMS & 14.90 & \textbf{13.53} & 0.12 & 5.17 & 4.03 & 0.21 \\ 
      +s.DNMS & \textbf{14.75} & 13.54 & \textbf{0.11} & \textbf{5.04} & \textbf{4.03} & \textbf{0.20} \\
      \bottomrule
  \end{tabular}
  }
\end{table}

In terms of mask ability, OBM still has a better performance compared with ROI prompt related models. 
In \cref{tab.f1score}, our model is compared with prompt-level LOFT, prompt-level Cascade LOFT, MTBR-Net(ICCV 2021)\cite{MTBRNet}, 
LOFT(TPAMI 2023)\cite{a12} and MLS-BRN(CVPR 2024)\cite{MLSBRN} in terms of the ultimate footprint masks. 
The first half of the form demonstrates our model outperforms all prompt-level models and the second half denotes our model is better than end-to-end models on BONAI dataset. 
Comparing models in different mode, easily we discover prompt-level models can give more accurate footprint segmentations, but we need to emphasize that this comparison has limitations.
Even for the same model LOFT, in prompt mode, LOFT will be told all potential buildings' location, and the f1score of prompt LOFT improved by 15.1\% compared with end-to-end LOFT.
The results are comprehensible: end-to-end outputs need to consider the problem of mistake detection and omitted buildings, while prompt-level models utilize inputted boxes ensure all buildings can get a relatively correct answer. 

\begin{table}[h]
  \centering
  \caption{F1score, Recall and Precision of finally predicted footprints on BONAI among different models\cite{a12}}
  \label{tab.f1score}
  
    \begin{tabular}{c|ccc}
    \toprule
    {\textbf{Model}}&F1score & Recall & Precision\\ \hline
    prompt LOFT &0.740 &0.853 &0.676 \\
    prompt Cas.LOFT &0.775 &\textbf{0.867} &0.718\\
    Ours(OBM) & \textbf{0.813} & 0.832 & \textbf{0.808}\\ \hline
    LOFT\cite{a12} & 0.643 & 0.653 & 0.634\\
    MTBR-Net\cite{MTBRNet} & 0.636 & 0.629 &0.643 \\
    MLS-BRN\cite{MLSBRN} & 0.664 & 0.668 & 0.659\\
    \bottomrule
    \end{tabular}

\end{table}

As shown in \cref{tab.1}, the roof regression of OBM is more accurate than LOFT and Cascade LOFT. In \cref{fig.5}, some results were visualized. 
Footprint quality is highly relied on roof quality and offset accuracy.

\subsection{Results on generalization test}
\begin{figure}[htbp]
	\centering
	\begin{minipage}{0.32\linewidth}
		\vspace{4pt}
		\centerline{\includegraphics[width=\textwidth]{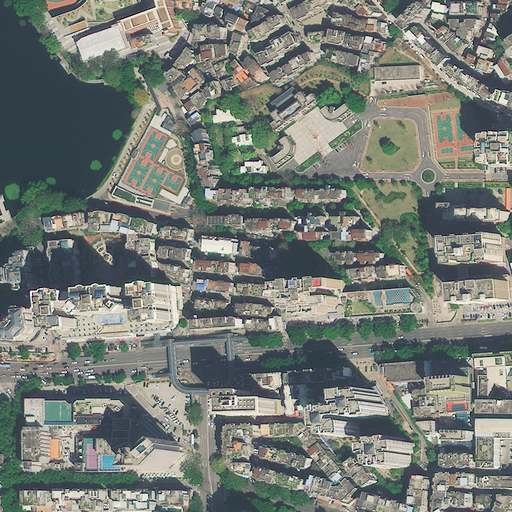}}
	\end{minipage}
	\begin{minipage}{0.32\linewidth}
		\vspace{4pt}
		\centerline{\includegraphics[width=\textwidth]{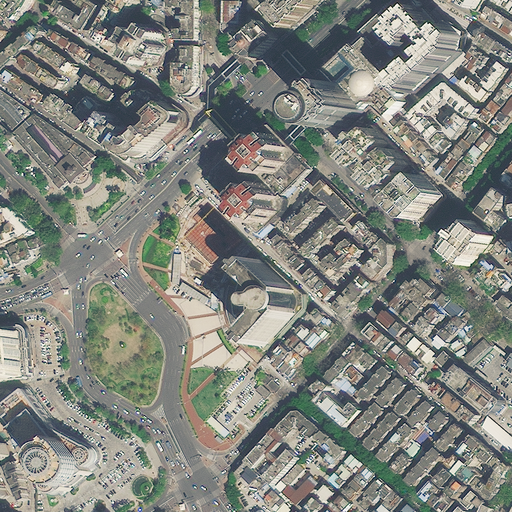}}
	\end{minipage}
	\begin{minipage}{0.32\linewidth}
		\vspace{4pt}
		\centerline{\includegraphics[width=\textwidth]{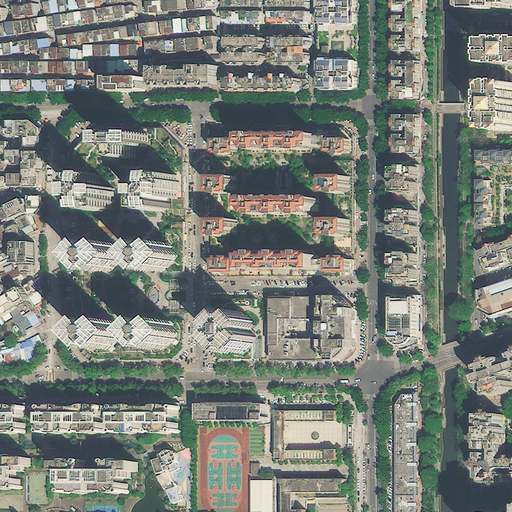}}
	\end{minipage}
	\begin{minipage}{0.32\linewidth}
		\vspace{4pt}
		\centerline{\includegraphics[width=\textwidth]{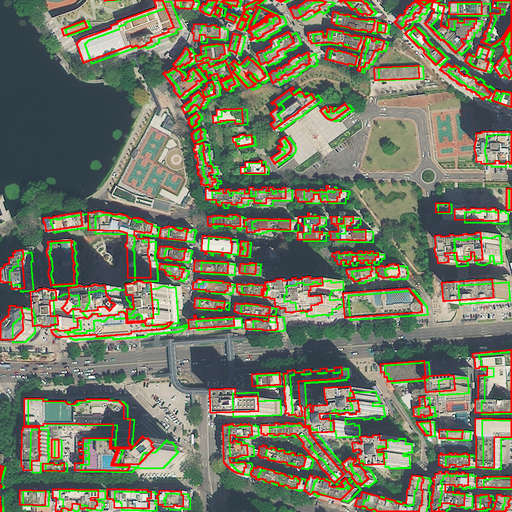}}
	\end{minipage}
	\begin{minipage}{0.32\linewidth}
		\vspace{4pt}
		\centerline{\includegraphics[width=\textwidth]{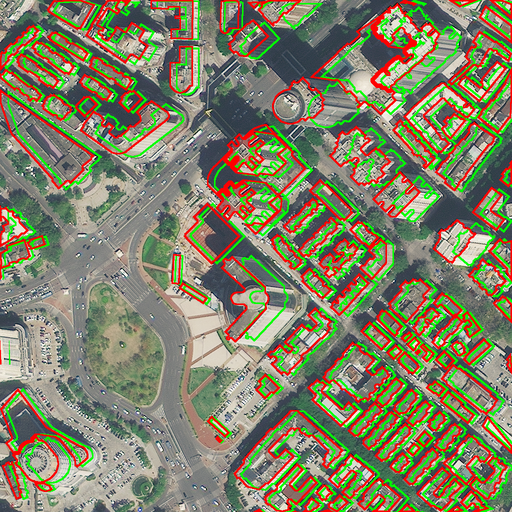}}
	\end{minipage}
	\begin{minipage}{0.32\linewidth}
		\vspace{4pt}
		\centerline{\includegraphics[width=\textwidth]{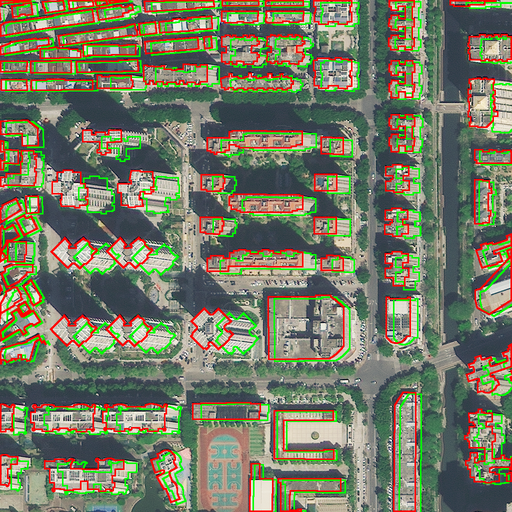}}
	\end{minipage}
	\caption{Some samples of Huizhou test dataset. In given pictures, red and green boundaries represent the edges of roof masks and footprint masks.}
  \label{fig.huizhou}
\end{figure}

In this part, we newly annotated a dataset with over 7,000 instances. The spatial resolution of images (shape 1024$\times$1024) is 0.5 m (same with BONAI\cite{a12}). 
All images are collected from Huizhou, China (Coordinates: 23.1125° N, 114.4155° E). Huizhou is a coastal city of Guangdong Province. 
Due to its advantageous geographical location, the reform and opening-up policy has rapidly transformed Huizhou into a modern city. 
This allowing Huizhou to have many complex buildings, including bungalows and skyscrapers. 
Within huizhou test set, 43.1\% of building offsets are longer than 10 pixels.
\cref{fig.huizhou} displayed some figures in this test dataset. 

\noindent 

\begin{table}[htbp!]
  \centering
  \caption{Detail results on newly annotated dataset, without extra training}
  \label{tab.22a}
  \resizebox{\linewidth}{!}{
  \begin{tabular}{c|rrr|rr}
  \toprule
      \textbf{Model} & \textbf{$aVL$} & \textbf{$aLL$} & \textbf{$aAL$} & IoU & BIoU \\ \hline
      p. LOFT & 10.40 & 9.009 & 0.770  \\ 
      +DNMS & \textbf{9.293} & \textbf{9.009} & \textbf{0.204} & 0.587 & 0.260\\ 
      +s.DNMS  & 9.475 & 9.077 & 0.268 \\ \hline
      p. Cas. LOFT & 10.82 & 8.964 & 0.844  \\ 
      +DNMS & \textbf{9.252} & \textbf{8.964} & \textbf{0.183} & 0.627 & 0.281\\
      +s.DNMS & 9.782 & 9.051 & 0.400 \\ \hline
      OBM (our) & 8.299 & 7.892 & 0.182  \\
      +DNMS & 8.233 & \textbf{7.891} & 0.167& \textbf{0.686} & \textbf{0.346} \\
      +s.DNMS & \textbf{8.229} & 7.892 & \textbf{0.161} \\ \bottomrule
  \end{tabular}
  }
\end{table}

\begin{figure*}[htbp]
\centering
\begin{minipage}{0.133\linewidth}
  \centerline{\includegraphics[width=\textwidth]{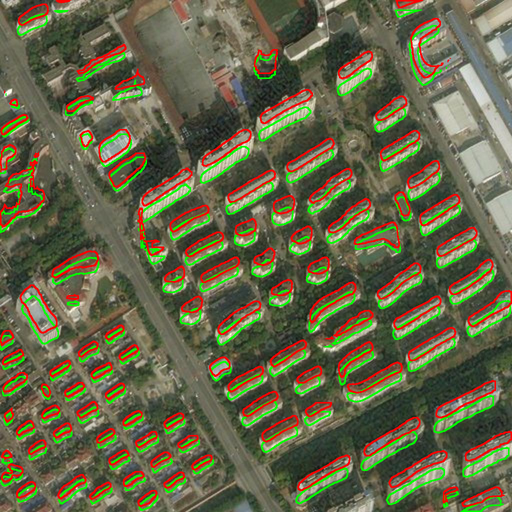}}
  \vspace{0.8pt}
  \centerline{\includegraphics[width=\textwidth]{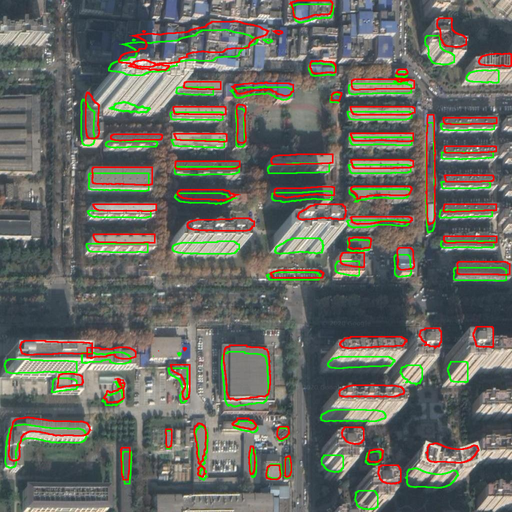}}
  \vspace{0.8pt}
  \centerline{\includegraphics[width=\textwidth]{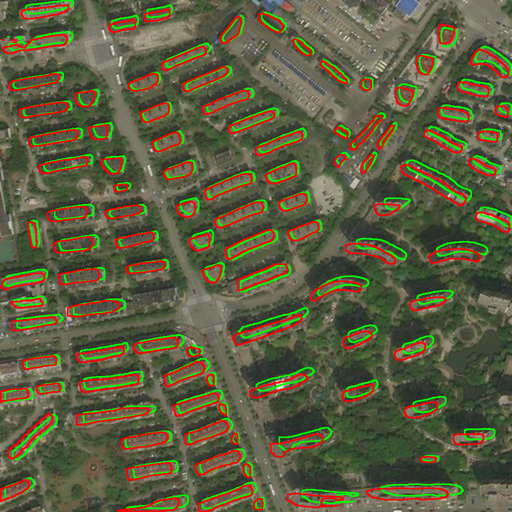}}
  \vspace{0.8pt}
  \centerline{\includegraphics[width=\textwidth]{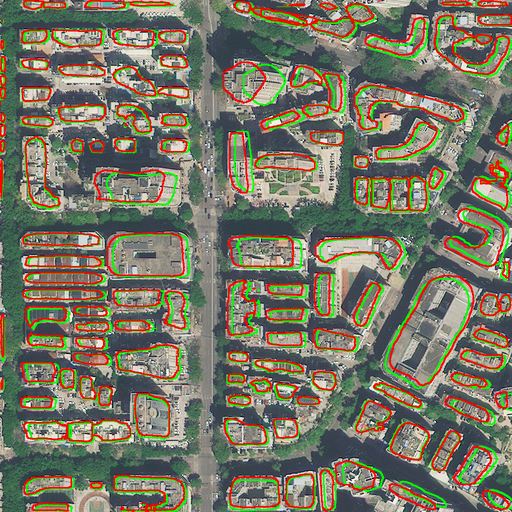}}
  \vspace{0.8pt}
  \centerline{\includegraphics[width=\textwidth]{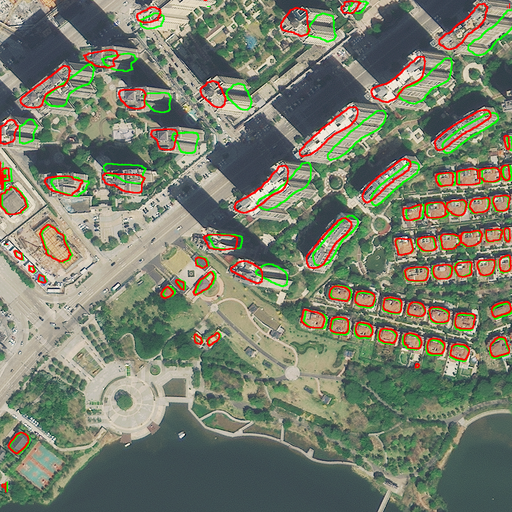}}
  \vspace{0.8pt}
  \centerline{\includegraphics[width=\textwidth]{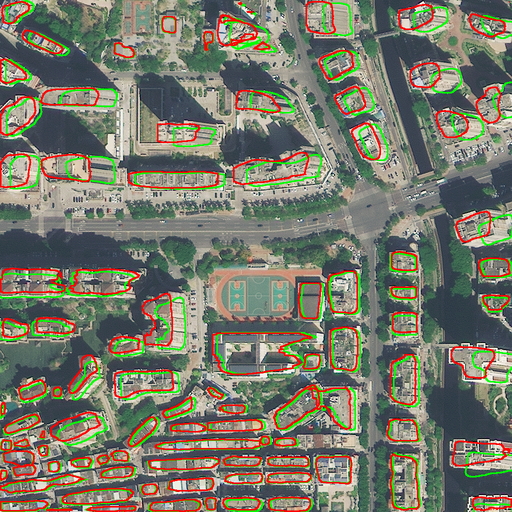}}
  \vspace{0.8pt}
  \centerline{\small{LOFT}}
\end{minipage}
\begin{minipage}{0.133\linewidth}
  \centerline{\includegraphics[width=\textwidth]{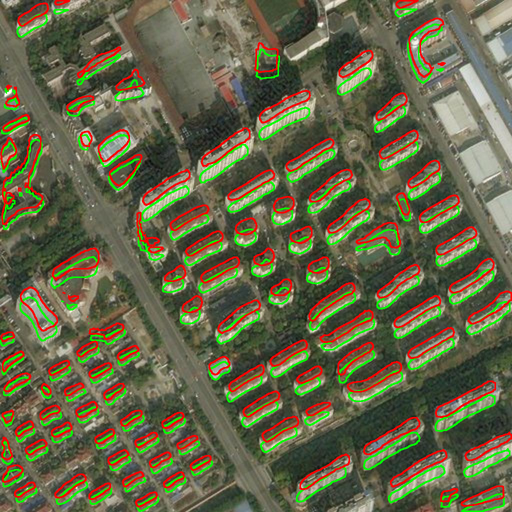}}
  \vspace{0.8pt}
  \centerline{\includegraphics[width=\textwidth]{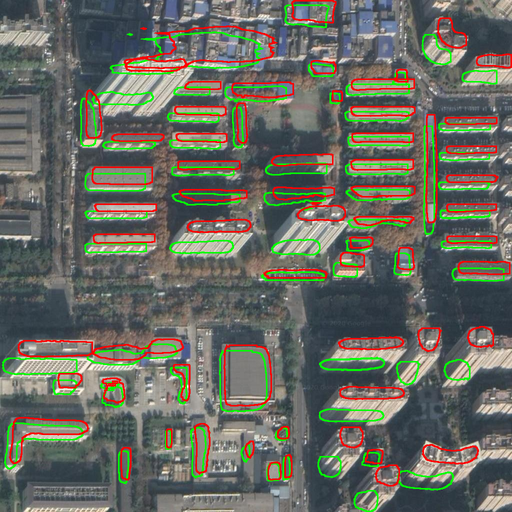}}
  \vspace{0.8pt}
  \centerline{\includegraphics[width=\textwidth]{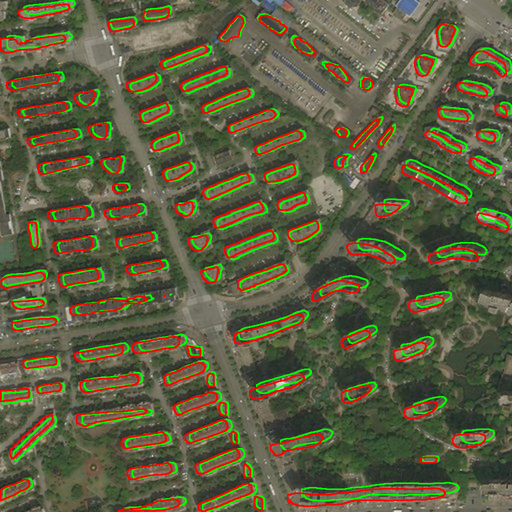}}
  \vspace{0.8pt}
  \centerline{\includegraphics[width=\textwidth]{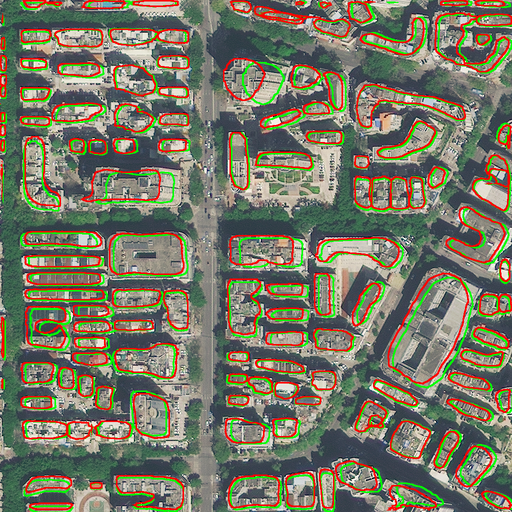}}
  \vspace{0.8pt}
  \centerline{\includegraphics[width=\textwidth]{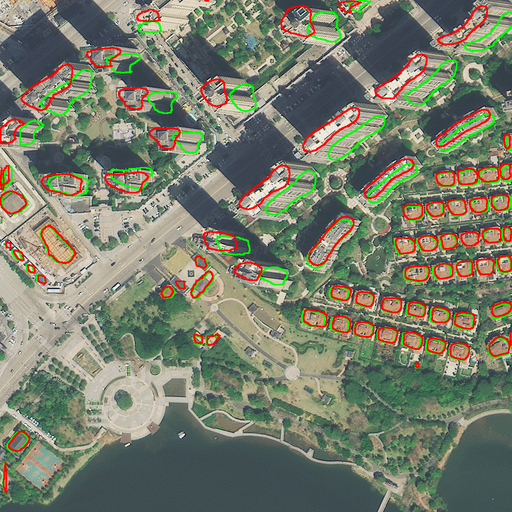}}
  \vspace{0.8pt}
  \centerline{\includegraphics[width=\textwidth]{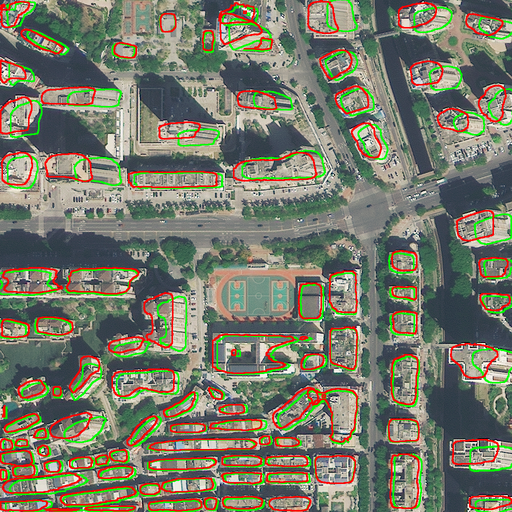}}
  \vspace{0.8pt}
  \centerline{\small{Cas. LOFT}}
\end{minipage}
\begin{minipage}{0.133\linewidth}
  \centerline{\includegraphics[width=\textwidth]{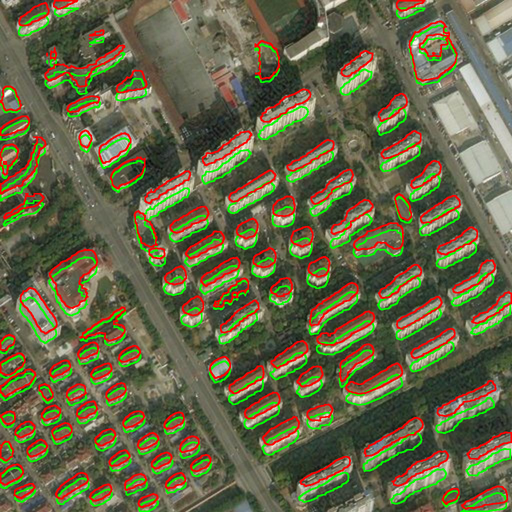}}
  \vspace{0.8pt}
  \centerline{\includegraphics[width=\textwidth]{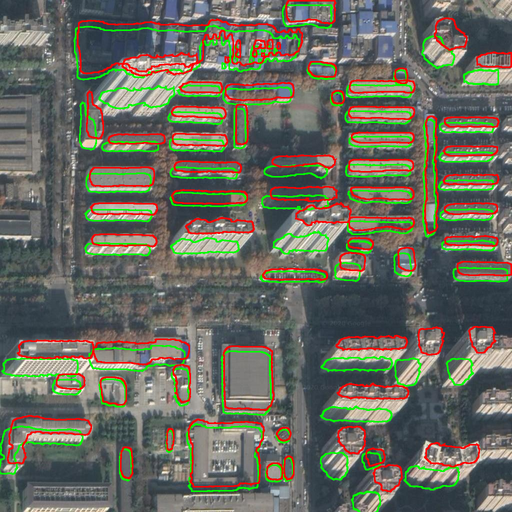}}
  \vspace{0.8pt}
  \centerline{\includegraphics[width=\textwidth]{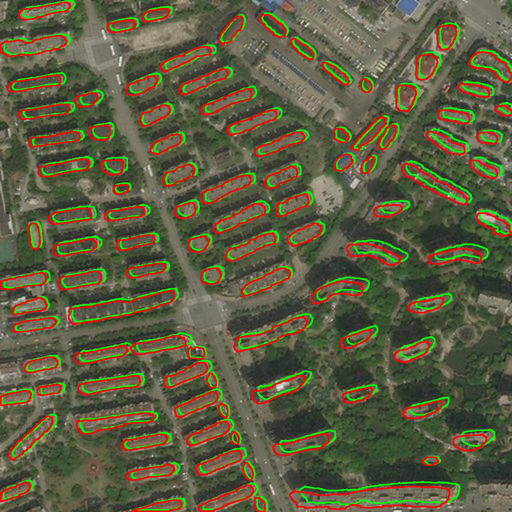}}
  \vspace{0.8pt}
  \centerline{\includegraphics[width=\textwidth]{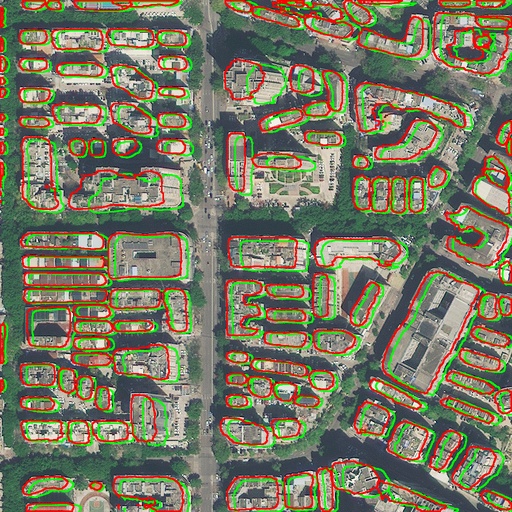}}
  \vspace{0.8pt}
  \centerline{\includegraphics[width=\textwidth]{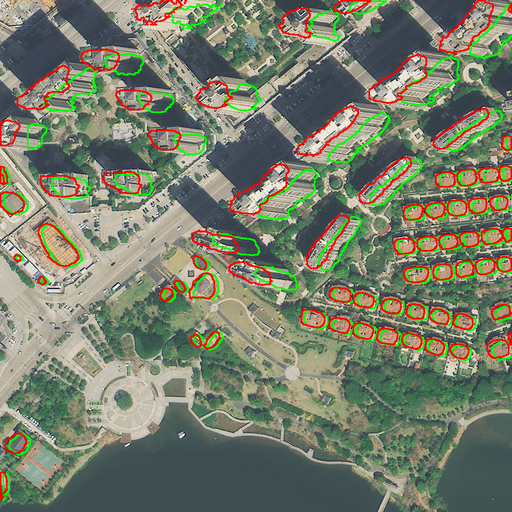}}
  \vspace{0.8pt}
  \centerline{\includegraphics[width=\textwidth]{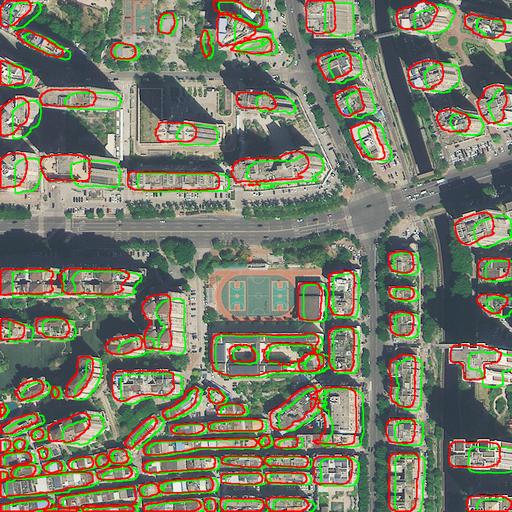}}
  \vspace{0.8pt}
  \centerline{\small{Ours}}
\end{minipage}
\begin{minipage}{0.133\linewidth}
  \centerline{\includegraphics[width=\textwidth]{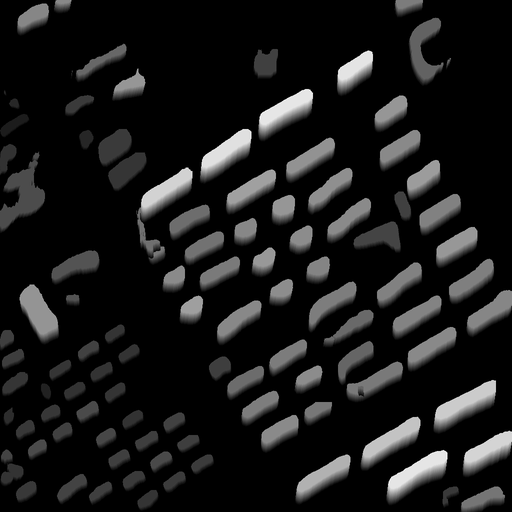}}
  \vspace{0.8pt}
  \centerline{\includegraphics[width=\textwidth]{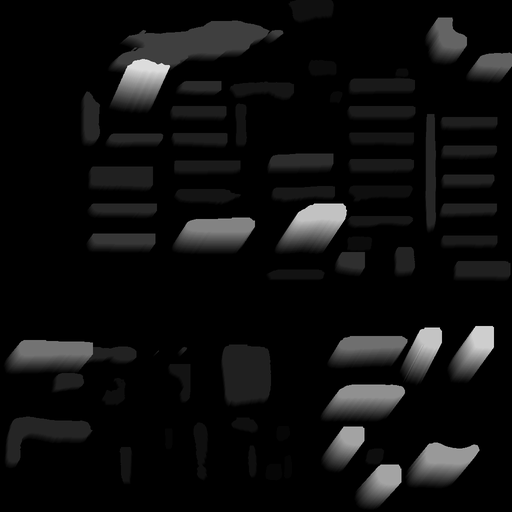}}
  \vspace{0.8pt}
  \centerline{\includegraphics[width=\textwidth]{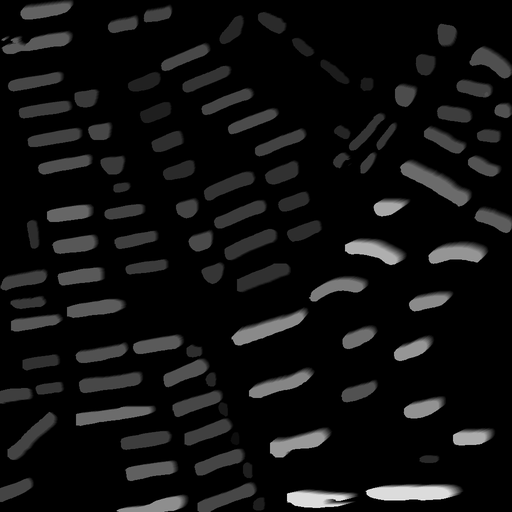}}
  \vspace{0.8pt}
  \centerline{\includegraphics[width=\textwidth]{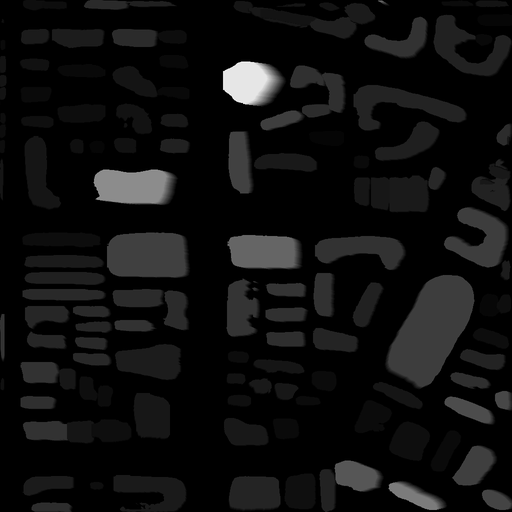}}
  \vspace{0.8pt}
  \centerline{\includegraphics[width=\textwidth]{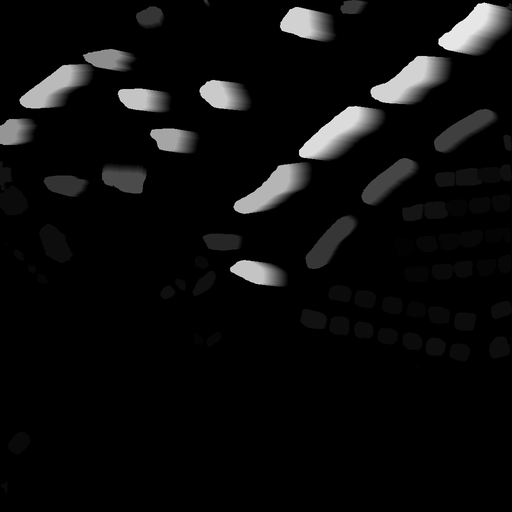}}
  \vspace{0.8pt}
  \centerline{\includegraphics[width=\textwidth]{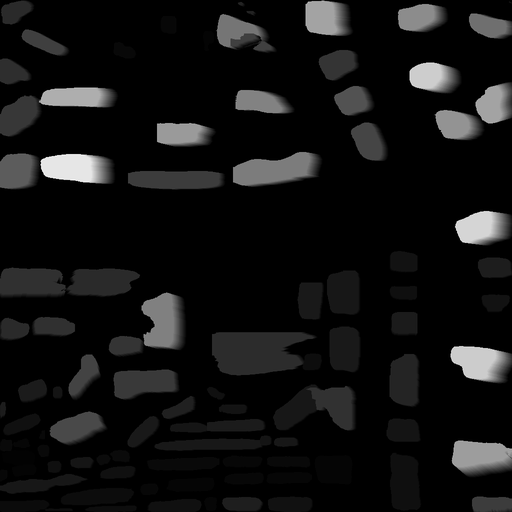}}
  \vspace{0.8pt}
  \centerline{\small{Height(LOFT)}}
\end{minipage}
\begin{minipage}{0.133\linewidth}
  \centerline{\includegraphics[width=\textwidth]{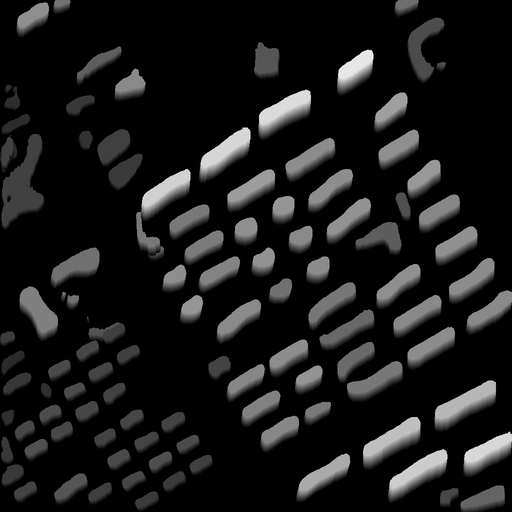}}
  \vspace{0.8pt}
  \centerline{\includegraphics[width=\textwidth]{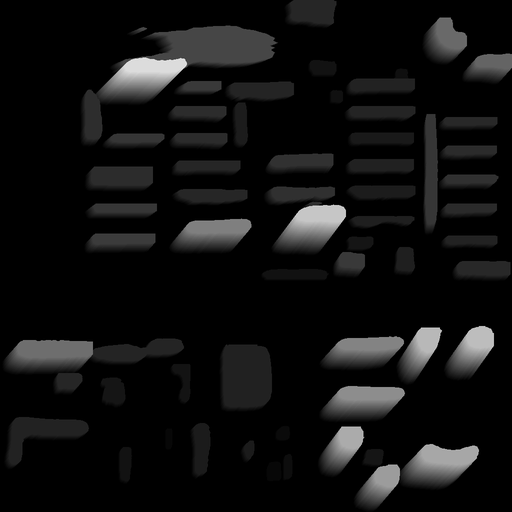}}
  \vspace{0.8pt}
  \centerline{\includegraphics[width=\textwidth]{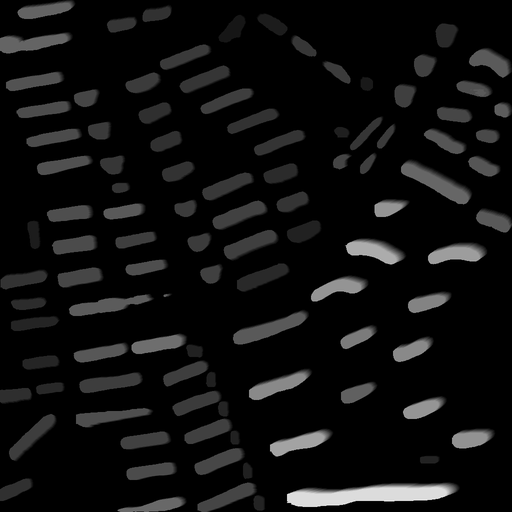}}
  \vspace{0.8pt}
  \centerline{\includegraphics[width=\textwidth]{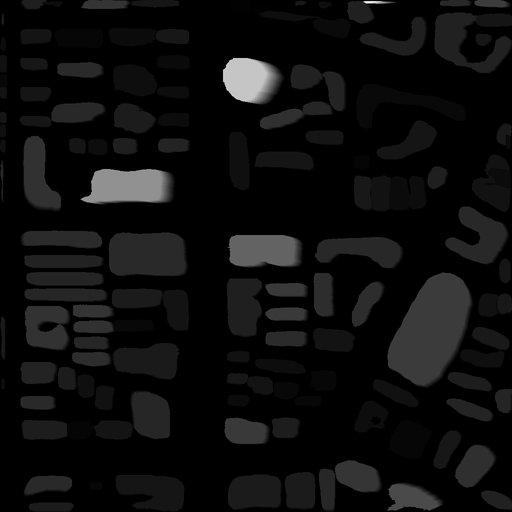}}
  \vspace{0.8pt}
  \centerline{\includegraphics[width=\textwidth]{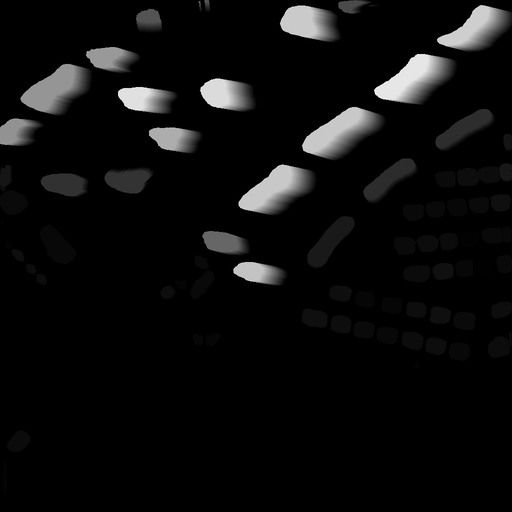}}
  \vspace{0.8pt}
  \centerline{\includegraphics[width=\textwidth]{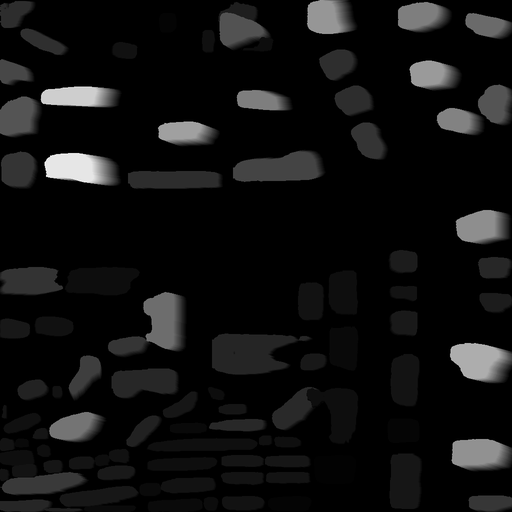}}
  \vspace{0.8pt}
  \centerline{\small{Height(Cas.LOFT)}}
\end{minipage}
\begin{minipage}{0.133\linewidth}
  \centerline{\includegraphics[width=\textwidth]{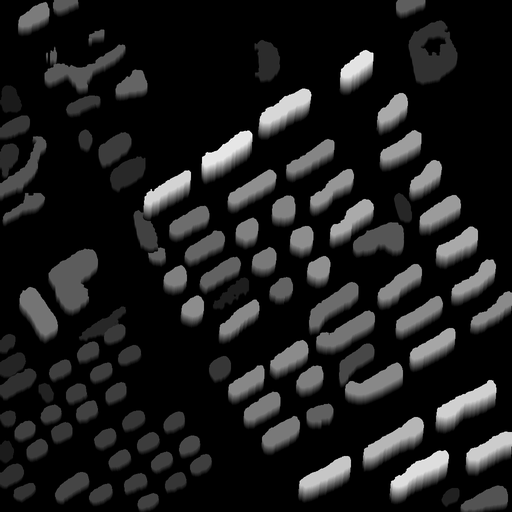}}
  \vspace{0.8pt}
  \centerline{\includegraphics[width=\textwidth]{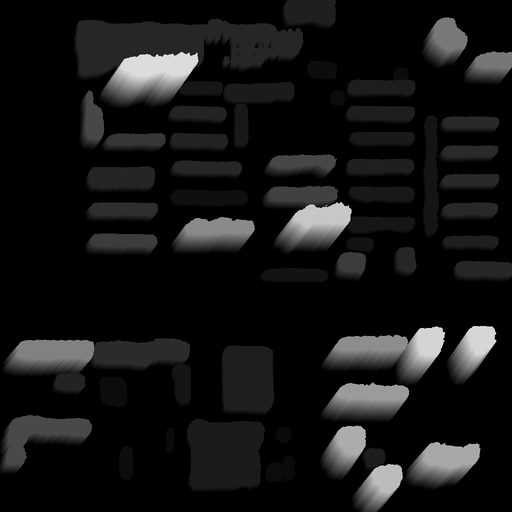}}
  \vspace{0.8pt}
  \centerline{\includegraphics[width=\textwidth]{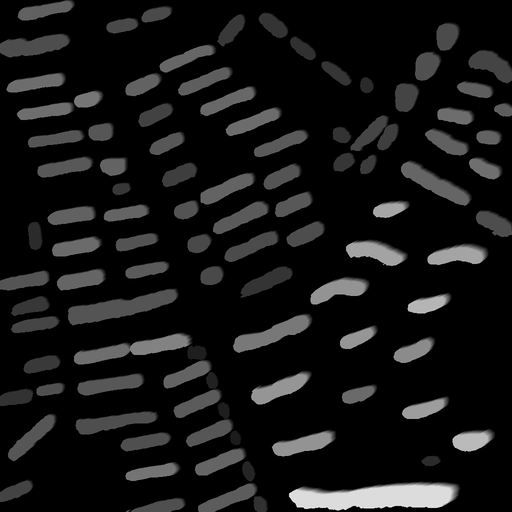}}
  \vspace{0.8pt}
  \centerline{\includegraphics[width=\textwidth]{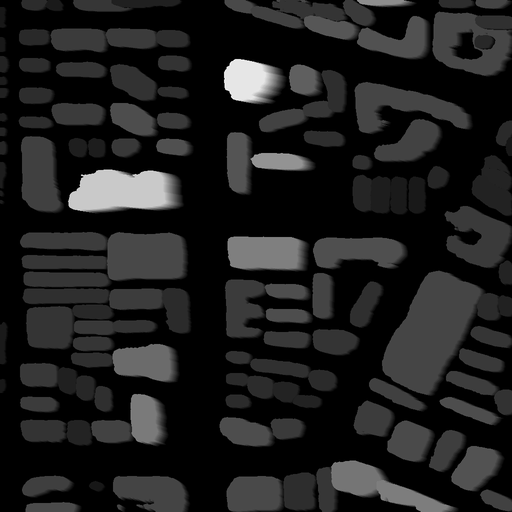}}
  \vspace{0.8pt}
  \centerline{\includegraphics[width=\textwidth]{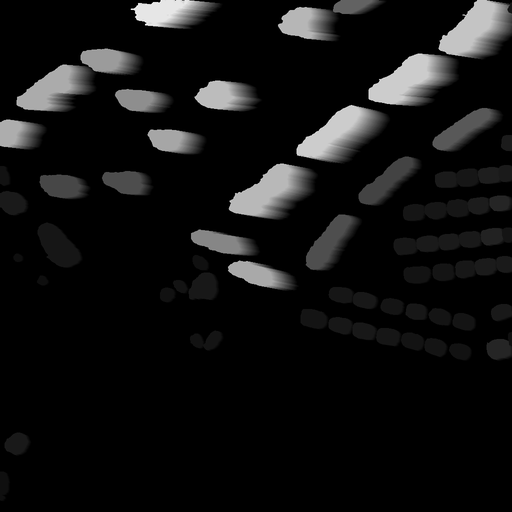}}
  \vspace{0.8pt}
  \centerline{\includegraphics[width=\textwidth]{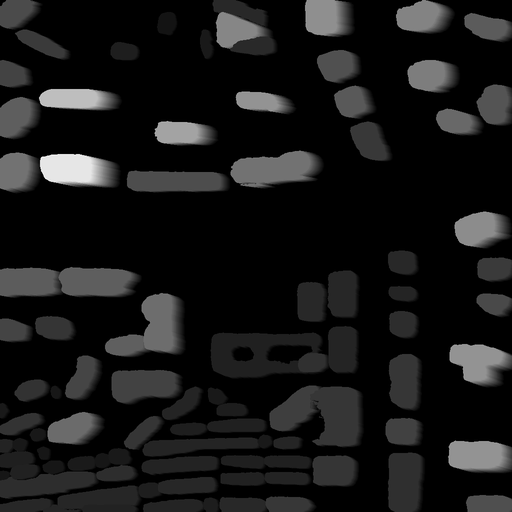}}
  \vspace{0.8pt}
  \centerline{\small{Height(Ours)}}
\end{minipage}
\begin{minipage}{0.133\linewidth}
  \centerline{\includegraphics[width=\textwidth]{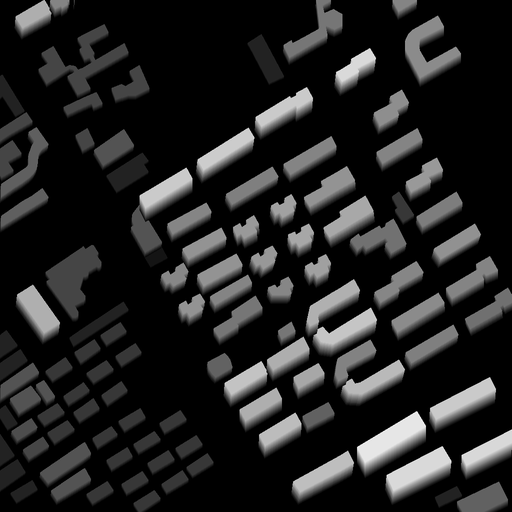}}
  \vspace{0.8pt}
  \centerline{\includegraphics[width=\textwidth]{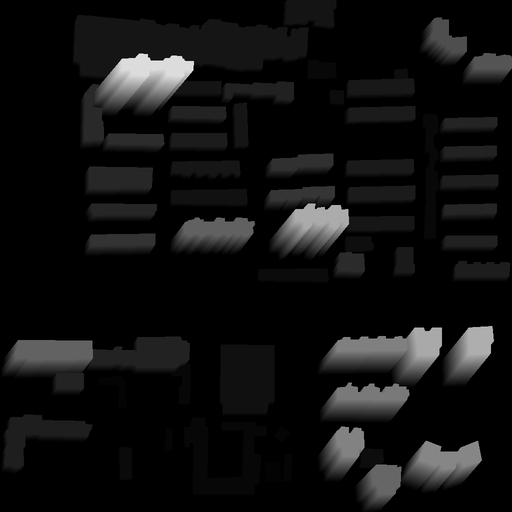}}
  \vspace{0.8pt}
  \centerline{\includegraphics[width=\textwidth]{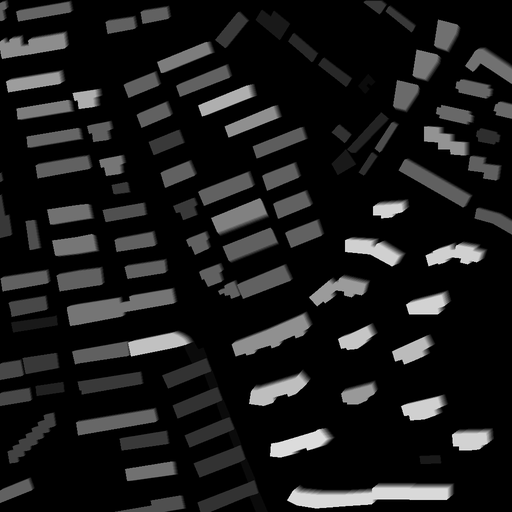}}
  \vspace{0.8pt}
  \centerline{\includegraphics[width=\textwidth]{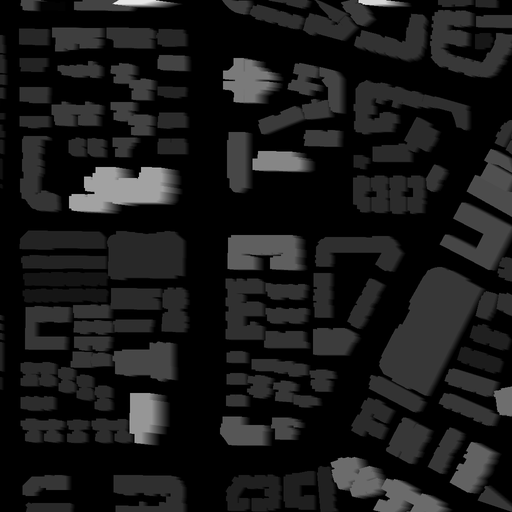}}
  \vspace{0.8pt}
  \centerline{\includegraphics[width=\textwidth]{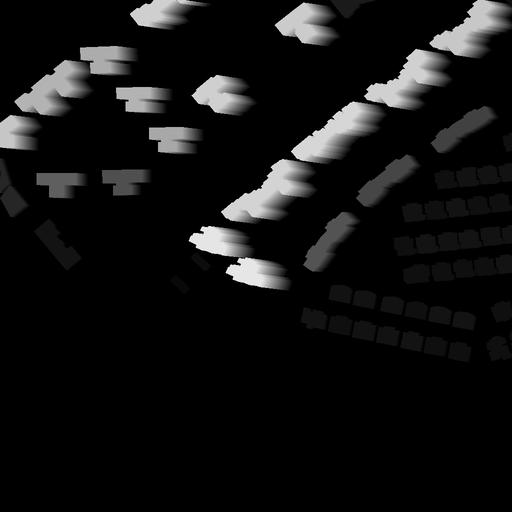}}
  \vspace{0.8pt}
  \centerline{\includegraphics[width=\textwidth]{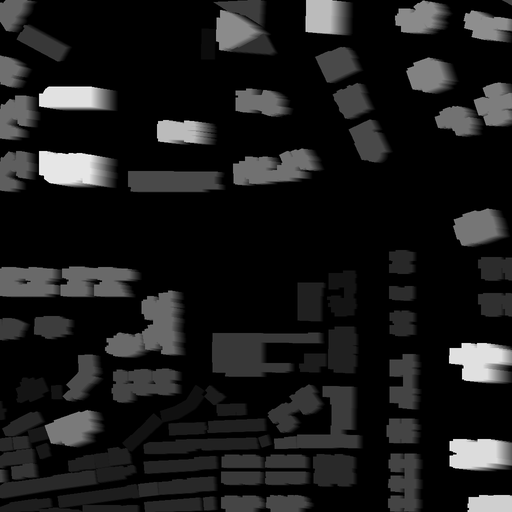}}
  \vspace{0.8pt}
  \centerline{\small{Ground Truth}}
\end{minipage}
\caption{We visualized inference results on test set. Former three images are from BONAI, the rest is generalization test on a newly annotated dataset. 
The relative height maps were generated by dragging and fading roof segmentation in the direction of offset, and the brightest building, valued 1, represents the highest one within the same image.}
\label{fig.5}
\end{figure*}

Then we conduct inference on this dataset without extra training. The length of offsets includes plenty of information \emph{e.g.} relative heights between each building. 
For better awareness of models, we visualized them in \cref{fig.5}.
Compared with former methods, our model has a 20.29\% lower vector loss and a 76.36\% lower angle loss in our brand-new dataset, 
shown in \cref{tab.22a}. Moreover, soft DNMS algorithm might not always provide better results than DNMS. 
This could attribute to the model performance: sometimes the direction of the longest offset is the best direction which has the minimal angle loss($AL$). 

To ensure the ability of our model, experiments were conducted on OmniCity-view3\cite{omnicity}, 
and related results can be found on \cref{tab.omnicity}. 
OmniCity-view3 is a dataset with offset annotations for 17,092 and 4,929 images (shape 512$\times$512) from train-val and test set respectively.
BONAI has 165,881 annotations for training and 66.8\% of them are longer than 10 pixels. 
In OmniCity-view3, there are 191,470 annotations, but 47.6\% of them are shorter than 10 pixels. 

In this experiment, we found a similar result with experiment on BONAI. 
Prompt-level test results widely outperform those end-to-end models. The reason may also be similar. 
Note that all the experimental models are trained on full OmniCity-view3 dataset. 
Our OBM enhances the F1-score by 5.7\% relative to prompt LOFT. The results indicate that our approach betters the \( aAL \) by 0.884 in comparison to prompt LOFT. 

\begin{table}[htbp!]
  \centering
  \caption{Results about offset and footprint quality on OmniCity-view3\cite{omnicity} dataset. 
  OBM* represents model trained on BONAI's zero-shot performance.}
  \label{tab.omnicity}
  \resizebox{\linewidth}{!}{
  \begin{tabular}{c|rrr|rrr}
  \toprule
      \textbf{Model} & \textbf{$aVL$} & \textbf{$aLL$} & \textbf{$aAL$} &Precision & Recall &F1score\\ \hline
      prompt LOFT & 7.575 & 5.287 & 0.699 & 0.729 & 0.846 & 0.770  \\ 
      prompt Cas.LOFT & 7.248 & 5.116&0.690 & 0.756 & \textbf{0.864} & 0.797\\ 
      OBM(Ours)  & \textbf{6.691} & \textbf{5.145} & \textbf{0.636}&\textbf{0.785}& 0.861&\textbf{0.814}\\ \hline
      OBM* & 9.622 & 7.214 & 1.274 &0.781&0.794& 0.781\\ \hline
      LOFT\cite{a12} & - & - & - &0.688&0.722&0.705\\
      MLS-BRN\cite{MLSBRN}&-&-&- & 0.696& 0.751  & 0.723 \\  \bottomrule
  \end{tabular}
  }
\end{table}

\cref{tab.omnicity_dnms} illustrates DNMS algorithms can improve the global performance on OmniCity-view3\cite{omnicity}. 
The improvements on LOFT and Cascade LOFT are very clear, \eg. soft DNMS can reduce $aVL$ by 0.348 pixels (4.6\%) and 0.369 pixels (5.1\%) respectively. 
This table also implies that OBM has a very stable ability to give out the right direction of building offsets, so the improvement caused by DNMS algorithms is limited but exists.

\begin{table}[htbp!]
  \centering
  \caption{Results about offset correction with DNMS algorithms on OmniCity-view3 dataset.}
  \label{tab.omnicity_dnms}
  \resizebox{\linewidth}{!}{
  \begin{tabular}{c|rrr|rrr}
  \toprule
      \textbf{Model} & \textbf{$mVL$} & \textbf{$mLL$} & \textbf{$mAL$} & \textbf{$aVL$} & \textbf{$aLL$} & \textbf{$aAL$} \\ \hline
      prompt LOFT & 54.31 & 48.53 & 0.652 & 7.575 & 5.287 & 0.699  \\ 
      +DNMS & 54.59 & \textbf{48.53} & 0.663 & 7.563 & \textbf{5.287} & 0.668  \\ 
      +soft DNMS& \textbf{54.14} & 48.66 & \textbf{0.623} & \textbf{7.227} & 5.301 & \textbf{0.604} \\ \hline

      prompt Cas.LOFT & 52.90 & 48.37&0.616 & 7.248 & 5.116 & 0.690\\ 
      +DNMS & 53.15 & \textbf{48.37}&0.641 & 7.216 & \textbf{5.116} & 0.641\\ 
      +soft DNMS & \textbf{52.84} & 48.48&\textbf{0.596} & \textbf{6.879} & 5.123 & \textbf{0.583}\\ \hline

      OBM(Ours)  & 56.90 & 53.67 & \textbf{0.659}&6.691& 5.145&0.636\\ 
      +DNMS  & 56.92 & \textbf{53.67} & 0.677&6.691& \textbf{5.145}&\textbf{0.623}\\
      +soft DNMS  & \textbf{56.89} & 53.68 & 0.662&\textbf{6.681}& 5.156&0.624\\ \bottomrule
  \end{tabular}
  }
\end{table}

On this dataset, $mVL$ and $mLL$ are abnormal compared with that of BOANI. 
A contributing factor for this was the imbalance of data distribution, which finally makes the mistake prediction of those longer offsets (over 100 pixels). 
$m\mathcal{L}$ reflects more on the comprehensive performance of the model at various lengths of the predicted offset, 
and $a\mathcal{L}$ reflects the performance among the whole dataset. 
In the training set of OmniCity, there are 365,168 instance annotations among 17,092 images, but only 1,735 of them (0.475\%)  longer than 100 pixels. 
This makes $VL_{(100,\infty)}$ reach 322.9 pixels, and finally leads to the abnormality of $mVL$. 
\section{Ablation}
\label{sec:ablation}

\subsection{Model performance influenced by training strategy}
LOFT and Cascade LOFT are originally designed for end-to-end instance segmentation.
To clarify whether training in prompt like OBM would influence convergence, 
we design extra experiments in \cref{tab.4}. From this table, we discover whether using RPN in training are a key component for final performance. 
\eg LOFT with RPN can lower $mVL$ for over 41.06\% compared with the LOFT trained without RPN. 
Apart from above, using different tasks to lead the train may also influence final results. 

\begin{table*}[htbp]
    \centering
    \caption{Ablation study of training methods. From this table, ROI based model needs to use RPN to reach better convergence, 
    while our OBM needs segmentation tasks to lead training.}
    \label{tab.4}
    \begin{tabular}{c|ccccc|cccccccc}
    \toprule
        \textbf{Model} & \textbf{RPN} & \textbf{Box} & \textbf{Roof} & \textbf{Buil.} & \textbf{Offset} & \textbf{$mVL$} & \textbf{$mLL$} & \textbf{$mAL$} & \textbf{$aVL$} & \textbf{$aLL$} & \textbf{$aAL$} \\ \hline
        ~ & ~ & ~ & ~ & ~ & \checkmark & 26.72 & 23.89 & 0.326 & 7.928 & 5.494 & 0.545 \\ 
        ~ & ~ & ~ & \checkmark & ~ & \checkmark & 27.74 & 24.78 & 0.325 & 8.003 & 5.568 & 0.525 \\
        LOFT & ~ & \checkmark & ~ & ~ & \checkmark & 20.62 & 17.56 & 0.268 & 7.431 & 5.049 & 0.522 \\ 
        ~ & ~ & \checkmark & \checkmark & ~ & \checkmark & 26.06 & 23.06 & 0.322 & 7.860 & 5.398 & 0.541 \\ 
        ~ & \checkmark & \checkmark & \checkmark& ~ & \checkmark & 15.36&12.59&0.180&6.117&4.507&0.318 \\ \hline
        Cas.LOFT & \checkmark & \checkmark & ~ & ~ & \checkmark & 17.63 & 15.15 & 0.208 & 6.514 & 4.889 & 0.336 \\
        ~ & \checkmark & \checkmark & \checkmark & ~ & \checkmark & 15.80 & 13.45 & 0.173 & 5.972 & 4.482 & 0.309 \\ \hline
        ~ & ~ & ~ & ~ & ~ & \checkmark & - & - & - & - & - & - \\ 
         Our& ~ & ~ & \checkmark & ~ & \checkmark & 15.34 & 14.01 & 0.121 & 5.221 & 4.103 & 0.232 \\ 
         OBM & ~ & ~ & ~ & \checkmark & \checkmark & 14.92 & 13.56 & 0.121 & 5.146 & 4.041 & 0.230 \\ 
        ~ & ~ & ~ & \checkmark & \checkmark & \checkmark & \textbf{14.84} & \textbf{13.53} & \textbf{0.117} & \textbf{5.083} & \textbf{4.027} & \textbf{0.220} \\ \bottomrule
    \end{tabular}

\end{table*}

Model-related information, including FPS and parameters, is provided in \cref{params}. FPS was computed on the same server with one RTX 3090. 
The size of OBM is small enough; however, transformer blocks featured with their lower efficiency. However, CNN based model can tolerate larger batch size while inferencing.  
Based on our experiments, the FPS of OBM was influences by the number of box prompts, while CNN based models perform more stably. 
For example, on BONAI dataset, the scale of images are larger than that of OmniCity, and the number of prompts are relatively larger. 
The speed of OBM was the slowest as shown in \cref{params}. But on OmniCity, OBM is the fastest model. 
\begin{table}[htbp]
    \centering
    \caption{This table illustrates the number of parameters in different model, and their computation effeciencies.}
    \label{params}

    \begin{tabular}{c|rr}
    \toprule
        \textbf{Model} & FPS & Params\\ \hline
        {Res50 + LOFT} & 3.20 & 77.70 M  \\ 
        {Res50 + Cas. LOFT} & 2.50 & 180.74 M \\ 
        {SAM-b + OBM} & 1.69 & 90.40 M  \\
\bottomrule
    \end{tabular}
    
\end{table}

\subsection{Structure test of OBM}
\label{sec:roam}

In \cref{tab.s1}, We also provide different variants of OBM. 
Surprisingly, we find all variants has a considerable improvement compared with former methods.

To further the understanding of ROAM structure, we provide testing results of OBM with different number of Adaptive Head. 
The baseline OBM is one Base Head (BoxCoder \( \alpha= 200\)) with three Adaptive Head (BoxCoder \( \alpha= 150,300,400\)).
In \cref{tab.roam}, we tested beseline OBM's Base Head with different combination of Adaptive Heads. 
Although a simple combination always gets a worse answer, three-heads-assisting system finally gave out the best answer. 
The model was also retrained with only Base Head, shown in (-, -, -), which may imply ROAM system can inhance the model's ability on offset predicting while training.
To ensure which was the contributing factor to the improvement, we have calculated the accuracy of length prediction for different heads. 
The accuracy illustrates smaller BoxCoder \( \alpha\) can better predict shorter offset, and the larger \( \alpha\) can provide length compensation in the prediction.  
Worth to mention, all heads can give out an almost-same-good offset direction. 
\begin{table}[htp]
    \centering
    \caption{Ablation study of ROAM on BONAI\cite{a12}. In the column of \textbf{Model}, 
    we use (1,0,0) to represent base head with the first adaptive head. At last, we retrain an OBM with no
    adaptive head, and use (-,-,-) to denote this experiment. }
    \label{tab.roam}
    \resizebox{\linewidth}{!}{
    \begin{tabular}{c|rrrrrr}
    \toprule
    \textbf{Model} & \textbf{$mVL$} & \textbf{$mLL$} & \textbf{$mAL$} & \textbf{$aVL$} & \textbf{$aLL$} & \textbf{$aAL$} \\ \hline
        (0,0,0) & 14.93&13.61&0.117&5.118&4.051&0.221\\ \hline
        (1,0,0) & 14.91 & 13.56 & 0.117 & 5.149 & 4.076 & 0.220 \\ 
        (0,1,0) & 24.76& 23.55 & 0.118&10.27&9.416&0.221 \\ 
        (0,0,1) & 39.09 & 38.11 & 0.118 & 15.01 & 14.26 & 0.222 \\ \hline
        (1,1,0) & 14.65 & 13.34 & \textbf{0.116} & 5.115 & 4.054 & \textbf{0.219} \\ 
        (1,0,1) & 14.23 & 13.48 & 0.116 & 5.978 & 4.941 & 0.219 \\ 
        (0,1,1) & 22.26 & 21.02 & 0.117 & 9.283 & 8.408 & 0.221 \\ \hline
        (1,1,1) & \textbf{14.84} & \textbf{13.53} & 0.117 & \textbf{5.083} & \textbf{4.027} & 0.220 \\
        (-, -, -) & 17.40 & 15.91 & 0.138 & 5.651 & 4.422 & 0.256\\
        \bottomrule
    \end{tabular}
    }
  \end{table}

\subsection{The robustness of the model to prompt errors}
\begin{figure*}[htbp]
	\begin{minipage}{0.33\linewidth}
		\vspace{1pt}
		\centerline{\includegraphics[width=\textwidth]{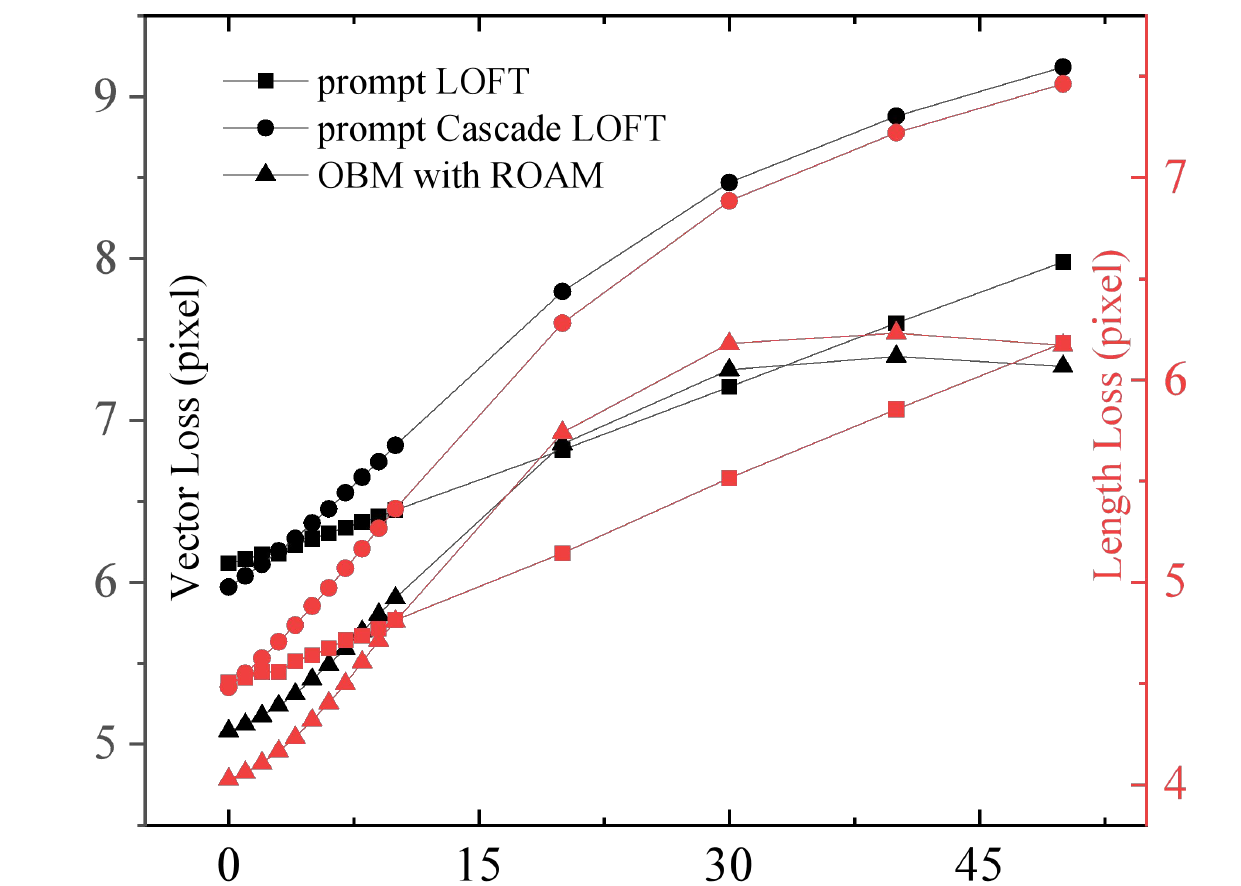}}
        \centerline{(a) $VL$ and $LL$ / Prompting Error}
	\end{minipage}
	\begin{minipage}{0.33\linewidth}
		\vspace{1pt}
		\centerline{\includegraphics[width=\textwidth]{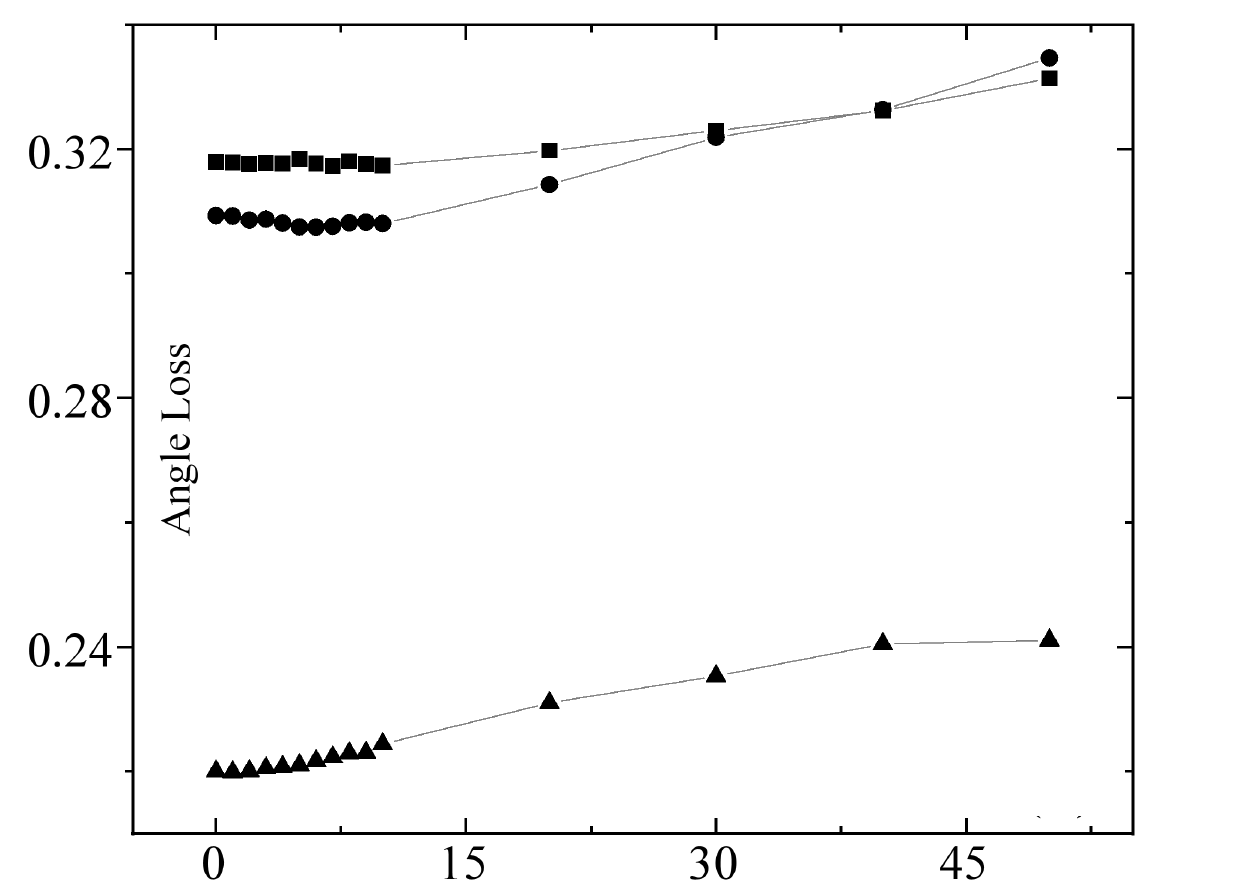}}
        \centerline{(b) $AL$ / Prompting Error}
	\end{minipage}
	\begin{minipage}{0.33\linewidth}
		\vspace{1pt}
		\centerline{\includegraphics[width=\textwidth]{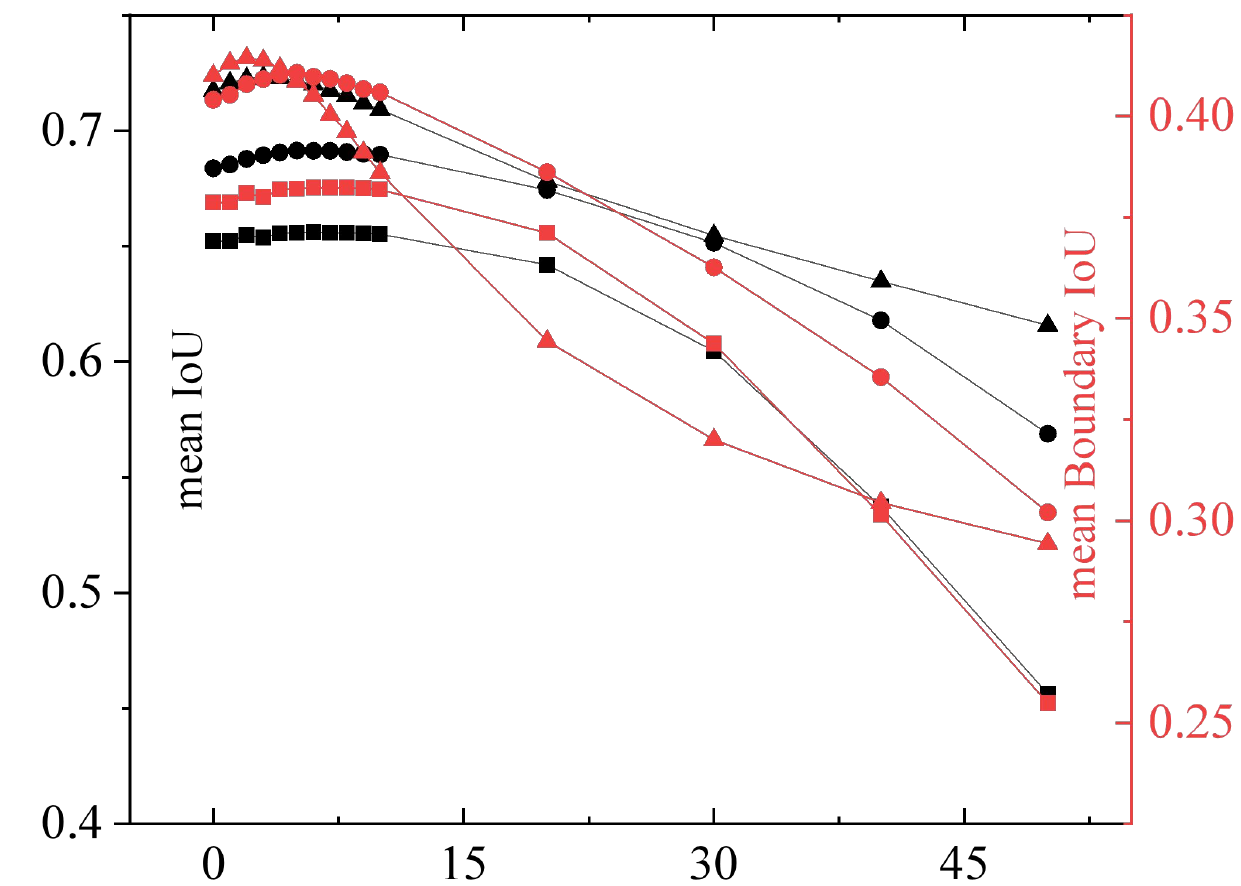}}
        \centerline{(c) Roof IoU and BIoU / Prompting Error}
	\end{minipage}
    \caption{The influence of prompt quality. 
    For each figure, $x$ axis represents the prompting error added on prompting boxes, and this random noise is measured by pixels. 
    In (a) and (b), we discover the quality drop of predictions mainly caused by the increase of length loss. 
    (c) tells us a suitable larger bounding box can improve the mask quality of prediction.}
	\label{fig.6}
\end{figure*}

As shown in \cref{fig.6}, 
given random noise still exists when human annotators drawing prompts, the performance of OBM was studied with prompt noise. 
We discover that noise within 10 pixels is controllable and even can slightly improve mask quality by 3\%. 
Performance of \cref{fig.6} could be explained by the same theory of \cref{whybetter}: 
when the input prompt box was noised, the valid features of ROI-based model will be noised, 
because the model will resize the patch features to the same size. 
However, attention based model can still focus on the valid features of those building with significant offsets and the building itself. 

\subsection{Different parameters in soft DNMS}
To tap into the potential of soft DNMS, we tested the influence of different $\alpha$, and the results are shown in \cref{tab.s2}.
and interactively use soft DNMS to improve offset quality. 
In conclusion, iteratively using soft DNMS has more significant improvement compared with adjusting $\alpha$.
\section{Discussion}

\subsection{Why does OBM perform better?}
\label{whybetter}
OBM performs better than LOFT and Cascade LOFT no matter in terms of roof regression or offset prediction. 
It is very common that people may attribute the improvement to the large-scale pre-trained backbone of OBM. 
Moreover, we want to point out that the improvement potentially attributes to the difference of how models get patch features. 
As shown in \cref{fig:3}, ROI based models will crop and resize features into a certain shape for all prompts. 
This shape often smaller than the original prompt size. An inevitable information loss was consequently caused during this process.  
Far from that, our transformer model gets those features more like using “soft roi-pooling” as discussed in \cite{a27}. 

\begin{figure}[htbp]
	\centering
	\begin{minipage}{0.24\linewidth}
		\vspace{4pt}
		\centerline{\includegraphics[width=\textwidth]{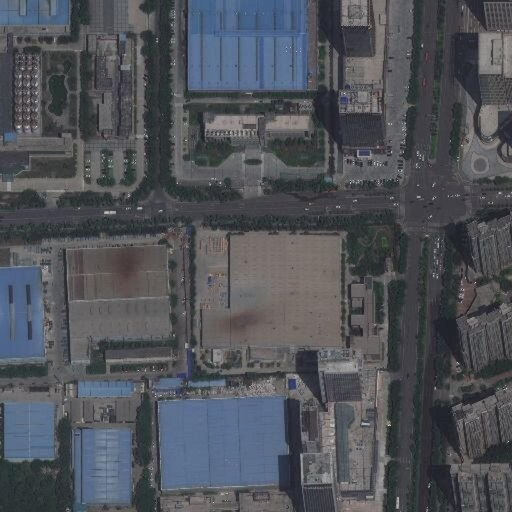}}
	\end{minipage}
	\begin{minipage}{0.24\linewidth}
		\vspace{4pt}
		\centerline{\includegraphics[width=\textwidth]{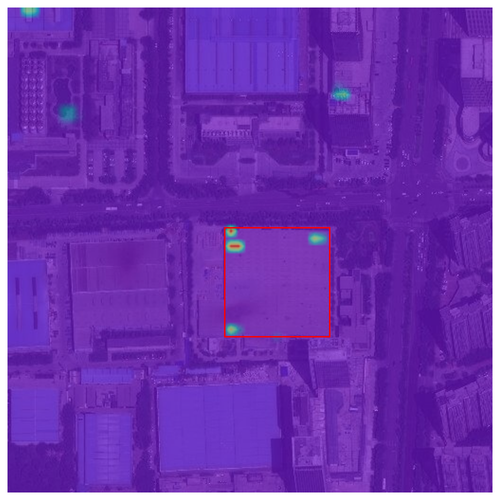}}
	\end{minipage}
	\begin{minipage}{0.24\linewidth}
		\vspace{4pt}
		\centerline{\includegraphics[width=\textwidth]{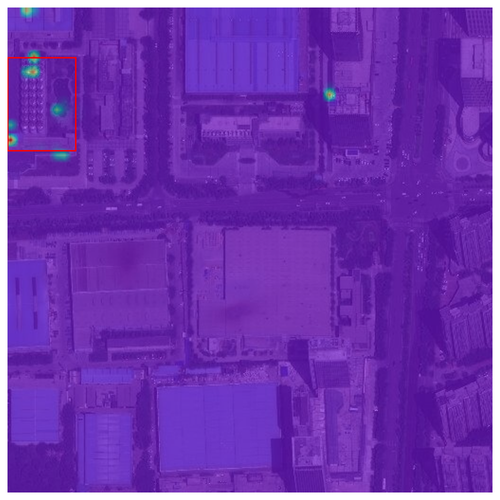}}
	\end{minipage}
    \begin{minipage}{0.24\linewidth}
		\vspace{4pt}
		\centerline{\includegraphics[width=\textwidth]{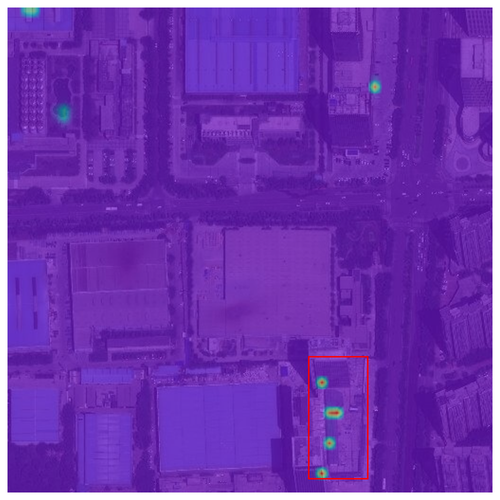}}
	\end{minipage}
	\caption{Offset tokens for red boxes will find buildings with significant offsets as references to get better performance, while ROI based methods only crop red box area. These results can be commonly found in experiments.}
  \label{fig.r1}
\end{figure}

Another key reason, we think, why OBM performs better on offset prediction is because of global attention.
We notice that in \cref{tab.1}, OBM has a better performance on those short offsets. We also discover this phenomenon in our generalization test.
OBM still remains it very powerful direction awareness. 
Far from only using features inside the prompt area like ROI based models, transformer structures in OBM still allow other parts of features involved while inferring. 
The final offset was not defined by a single area, but by several related areas. 
In other words, although shorter offsets are more difficult to predict, OBM obtains extra information from other buildings via attention. 
We visualized the offset tokens in \cref{fig.r1} to show this phenomenon.  
\noindent 

\subsection{Different structures of OBM}
\label{sec:obm_structure}
To ensure the best structure of our model. We designed four structures of OBM in \cref{tab.s1}. 
D. D. is a double decoder to predict offsets, roof and building separately. 
Share q. means different heads in ROAM using a shared offset token.

We tested all kinds of potential training methods to find the best performance of mentioned model the results are as \cref{tab.s1}. 
The final row is our final OBM.
\begin{table}[!htp]
    \centering
    \caption{Different structure of our model trained on BOANI\cite{a12}}
    \label{tab.s1}
    \resizebox{\linewidth}{!}{
    \begin{tabular}{c|rrrrrr}
    \toprule
    \textbf{Model} & $mVL$ & $mLL$ & $mAL$ & $aVL$ & $aLL$ & $aAL$ \\ \hline
        D. D. & 14.98 & 13.65 & 0.12& 5.20 & 4.13 & 0.23 \\ 
        Share q. & 18.53 & 16.89 & 0.15 & 5.83 & 4.66 & 0.24 \\ \hline
        OBM & \textbf{14.84} & \textbf{13.53} & \textbf{0.12} & \textbf{5.08} & \textbf{4.03} & \textbf{0.22} \\ 
        \bottomrule
    \end{tabular}
    }
  \end{table}

\subsection{Why does DNMS work?}
DNMS algorithms are designed to improve offset quality by correcting offset angle.
In \cref{tab.1}, easily we find shorter offsets often have poor performance while longer offsets were predicted well in terms of their direction. 
Naturally, we rise an idea: we can use those longer prediction to correct those shorter. DNMS algorithms will not delete subjects but only improve offset angle.

After experiments, DNMS algorithms can improve offset quality by correct the angle of offsets especially for shorter offsets. 
This is proved by \cref{tab.22b} and \cref{tab.d}.
\begin{table}[htbp]
    \centering
    \caption{In this table, DNMS and soft DNMS correct the angle for those shorter offsets. The circumstances of OBM are similar.}
    \label{tab.d}
    \resizebox{\linewidth}{!}{
    \begin{tabular}{c|rrr}
    \toprule
        \textbf{Model} & \textbf{\( AL_{(0,10)} \)} & \textbf{\( AL_{(10,20)} \)} & \textbf{\( AL_{(20,30)} \)} \\ \hline
        {LOFT} & 0.546 & 0.205 & 0.170 \\ 
        {LOFT+DNMS} & 0.371 & 0.179 & 0.164 \\ 
        {LOFT+s.DNMS} & 0.380 & 0.158 & 0.152 \\ \hline
        {Cas. LOFT} & 0.549 & 0.190 & 0.148 \\ 
        {Cas. LOFT+DNMS} & 0.381 & 0.177 & 0.161 \\ 
        {Cas.LOFT+s.DNMS} & 0.391 & 0.150 & 0.132 \\ \hline
        {OBM} & 0.365 & 0.152 & 0.137 \\
        {OBM+DNMS} & 0.351 & 0.161 & 0.143 \\
        {OBM+s.DNMS} & 0.341 & 0.146 & 0.135 \\ \bottomrule
    \end{tabular}
    }
\end{table}

\subsection{How to use soft DNMS for better?}
\label{sec:ssdnms}
DNMS once used, the angles of offsets in the whole image will be unified compulsorily. 
Soft DNMS is a soft version of DNMS. We need to determine the parameter $\alpha$ in soft DNMS to get the best performance. 
We use the prediction of LOFT as an example to show the effect of $\alpha$ in \cref{tab.s2}.
\begin{table}[!htp]
    \centering
    \caption{Different {$\alpha$} of soft DNMS test on BOANI\cite{a12} by LOFT}
    \label{tab.s2}
    \resizebox{\linewidth}{!}{
      \begin{tabular}{c|rrrrrr}
      \toprule
      \textbf{{$\alpha$}} & \textbf{$mVL$} & \textbf{$mLL$} & \textbf{$mAL$} & \textbf{$aVL$} & \textbf{$aLL$} & \textbf{$aAL$} \\ \hline
          -0.3&14.75&12.61&0.1380&5.714&4.506&0.2419\\
          -0.2&14.76&12.61&0.1371&5.706&4.506&0.2387\\
          -0.1&14.77&12.61&0.1363&5.699&4.506&0.2359\\\hline
          0   &14.78&12.62&0.1360&5.694&4.505&0.2339\\ 
          0.1 &14.79&12.62&\textbf{0.1358}&5.690&4.503&0.2321\\ 
          0.2 &14.80&12.62&0.1359&5.689&4.502&0.2311\\ 
          0.3 &14.81&12.62&0.1364&5.689&4.500&\textbf{0.2307}\\ 
          0.4 &14.82&12.62&0.1372&5.691&4.499&0.2308\\
          0.5 &14.84&12.62&0.1378&5.694&4.498&0.2316\\
          0.6 &14.84&12.61&0.1384&5.699&4.496&0.2333\\
          0.7 &14.86&12.61&0.1391&5.706&4.494&0.2353\\
          0.8 &14.86&12.61&0.1399&5.714&4.492&0.2379\\
          0.9 &14.87&12.61&0.1407&5.723&4.490&0.2404\\
          1.0 &14.88&12.61&0.1417&5.733&4.488&0.2434\\
          \hline
          LOFT&15.36&12.59&0.1803&6.117&4.507&0.3179\\
          \bottomrule
      \end{tabular}
    }

\end{table}

As our DNMS algorithms do not delete any offsets, 
it is also possible to iteratively use soft DNMS. 
We use the prediction of LOFT as an example to show the effect of iterative soft DNMS in \cref{tab.s3}.
\begin{table}[!htp]
    \centering
    \caption{Different iterations test on soft DNMS}
    \label{tab.s3}
    \resizebox{\linewidth}{!}{
      \begin{tabular}{c|rrrrrr}
      \toprule
      \textbf{{Iteration}} & \textbf{$mVL$} & \textbf{$mLL$} & \textbf{$mAL$} & \textbf{$aVL$} & \textbf{$aLL$} & \textbf{$aAL$} \\ \hline
          1 &14.81&12.62&0.1364&5.689&4.500&0.2307\\ 
          2 &14.72&12.63&0.1316&5.649&4.501&0.2221\\
          3 &14.67&12.63&0.1297&5.634&4.500&0.2190\\
          4 &14.64&12.62&0.1284&5.625&4.500&0.2173\\
          5 &14.62&12.62&0.1275&5.620&4.500&0.2163\\
          6 &14.62&12.63&0.1277&5.620&4.500&0.2164\\
          7 &14.62&12.63&0.1273&5.617&4.499&0.2159\\
          8 &14.61&12.63&0.1270&5.617&4.499&0.2157\\
          9 &14.60&12.63&0.1269&\textbf{5.617}&4.499&\textbf{0.2157}\\
      {$\infty$}&\textbf{14.56}&12.64&\textbf{0.1258}&5.622&\textbf{4.499}&0.2165\\
          \hline
          LOFT&15.36&\textbf{12.59}&0.1803&6.117&4.507&0.3179\\
          \bottomrule
      \end{tabular}
    }
  \end{table}

\subsection{Decoupling model's benefits}
Our OBM model has better performance than LOFT and Cascade LOFT, which is more reliable and robust.
Powerful prompt encoder makes sure that OBM can integrate almost any other models, \emph{e.g.} building detection, roof semantic segmentation and so on, to provide prompts. 
That means in real-world applications, data producers can get better results, even using the same RPN as \cite{a12} to provide prompts. 
For those lost target, data annotators can also label them out in a less time-consuming way because basically we can provide a roof segmentation at least. 

Moreover, our prompt-level metrics for offsets are fairer and more accurate to display abilities of model. 
That is because regressing offsets and regressing instances are very different. 
Once offsets are predicted in instance way, the one-to-many predicted offsets will dilute the final result. 
That is the reason why Wang\cite{a12} measured LOFT with only 4.7 pixels loss, while we have 6.12 pixels. 

In the past, COCO metrics, F1score, Precision and Recall were widely used in measure footprint masks\cite{a12,omnicity,MLSBRN,MTBRNet};
however, these metrics today may not suitable in measuring prompt-level models. As we saw in \cref{tab.f1score} and \cref{tab.omnicity}, 
prompt-level models easily defeat those end-to-end models. 
The reasons are as follows:
Metrics, such as F1score, Precision and Recall, indicate models' ability of predicting an output without any references. 
  However, the proposal network usually perform unstable: smaller IoU threshold will give out low quality instances whose masks are insufficient to cover the related items, 
  while bounding boxes with higher IoU threshold are rare in one predicted output. In other words, end-to-end models tend to omit some important samples or give out mistake outputs. 
Consequently, prompt-level models avoid aforementioned problems by directly import valid bounding boxes to activate boxes regions. 
In other words, metrics for end-to-end models inflect the capability of finding right and good footprints, and metrics for prompt-level models will emphasize more on whether the model can give out high quality masks for each building. 

\subsection{Potential improvements}
OBM inherits from SAM, whose final results are up-sampled by 16$\times$. SAM-HQ\cite{a17} improved boundary quality and mask IoU by resampling output features of SAM's backbone.
Similarly, we can also improve OBM by the same way. 

For buildings with significant roof-to-footprint offsets (exceeding 100 pixels), the models consistently struggles to provide satisfactory answers. Because this type of building is relatively scarce in the training dataset.

On the other hand, based on our understanding, extracting building roof, footprint, building body, and the offset in very-high-resolution remote sensing image is a problem which can be defined by each two of four. 
\emph{e.g.}, if we can find both segmentation of building roof and footprint, we can consequently calculate the offset. 
Similarly, if we can find the building-body segmentation and roof segmentation, we can also calculate the offset and footprint.
Based on this idea, there are many other aspects we can improve the predicting process. 

\section{Conclusion}
In this paper, we propose a decoupling Offset-Building Model (OBM) for Building Footprint Extraction (BFE) problem. 
This model extracts building footprints in prompt-level by extract a roof segmentation and roof-to-footprint offset. 
OBM demonstrated outstanding performance. 
We discovered that buildings with significant offsets in images are more likely to be extracted correctly, especially in terms of offset direction.
Moreover, integrating common pattern of building offset extraction with NMS algorithm, DNMS algorithms leverage those longer offsets with better direction to improve global performance. 
DNMS algorithm replaces directions of all offsets by the direction of the longest offset, 
while soft DNMS algorithm adjusts the directions referring all offsets.
Delicate roof semantic segmentation and accurate offset recognition allow OBM to generate relative height map of buildings in the same local area.
For further evaluation of OBM and DNMS algorithms, a newly annotated dataset was launched to test model generalization, 
and OBM and DNMS algorithms gave out a reliable result. 
Although OBM inherits the semantic ability of SAM, the shortage of SAM also exists in OBM. 
In the future, we will focus on how to get building masks whose boundaries are more closely aligned with buildings. 
\bibliographystyle{IEEEtran}
\bibliography{main} 

\begin{thebibliography}{10}
\providecommand{\url}[1]{#1}
\csname url@samestyle\endcsname
\providecommand{\newblock}{\relax}
\providecommand{\bibinfo}[2]{#2}
\providecommand{\BIBentrySTDinterwordspacing}{\spaceskip=0pt\relax}
\providecommand{\BIBentryALTinterwordstretchfactor}{4}
\providecommand{\BIBentryALTinterwordspacing}{\spaceskip=\fontdimen2\font plus
\BIBentryALTinterwordstretchfactor\fontdimen3\font minus \fontdimen4\font\relax}
\providecommand{\BIBforeignlanguage}[2]{{%
\expandafter\ifx\csname l@#1\endcsname\relax
\typeout{** WARNING: IEEEtran.bst: No hyphenation pattern has been}%
\typeout{** loaded for the language `#1'. Using the pattern for}%
\typeout{** the default language instead.}%
\else
\language=\csname l@#1\endcsname
\fi
#2}}
\providecommand{\BIBdecl}{\relax}
\BIBdecl

\bibitem{a1}
F.~Zhang, N.~Nauata, and Y.~Furukawa, ``Conv-{{MPN}}: {{Convolutional Message Passing Neural Network}} for {{Structured Outdoor Architecture Reconstruction}},'' in \emph{IEEE Conf. Comput. Vis. Pattern Recog.}, 2020, pp. 2795--2804.

\bibitem{a2}
Y.~T. Debao~Huang and R.~Qin, ``An evaluation of planetscope images for 3d reconstruction and change detection - experimental validations with case studies,'' \emph{GIScience and Remote Sensing}, vol.~59, pp. 744--761, 2022.

\bibitem{a3}
J.~Mahmud, {\relax TRUE}.~Price, A.~Bapat, and J.-M. Frahm, ``Boundary-{{Aware 3D Building Reconstruction From}} a {{Single Overhead Image}},'' in \emph{IEEE Conf. Comput. Vis. Pattern Recog.}, 2020, pp. 438--448.

\bibitem{a4}
J.~Yuan, ``Learning {{Building Extraction}} in {{Aerial Scenes}} with {{Convolutional Networks}},'' \emph{IEEE Trans. Pattern Anal. Mach. Intell.}, vol.~40, no.~11, pp. 2793--2798, 2018.

\bibitem{a5}
S.~Chen, Y.~Shi, Z.~Xiong, and X.~X. Zhu, ``Htc-dc net: Monocular height estimation from single remote sensing images,'' \emph{{IEEE} Trans. Geosci. Remote Sens.}, vol.~61, pp. 1--18, 2023.

\bibitem{landprice}
A.~Hou, J.~Liu, Y.~Tao, S.~Jiang, K.~Li, Z.~Zheng, J.~Xia, Y.~He, M.~Zhu, G.~Zhou, H.~Zhang, and J.~Li, ``Land price assesment based on deep neural network,'' in \emph{{IEEE} Int. Geosci.and Remote Sens. Symp.}, 2019, pp. 3053--3056.

\bibitem{urbanregion}
F.~Mou, R.~Kong, K.~Li, Z.~Zheng, J.~Xia, Y.~He, M.~Zhu, G.~Zhou, H.~Zhang, Z.~Liu, A.~Hou, L.~Jiang, S.~Wang, and J.~Li, ``Urban functional regions discovering based on deep learning,'' in \emph{{IEEE} Int. Geosci.and Remote Sens. Symp.}, 2019, pp. 9462--9465.

\bibitem{huang2022evaluation}
D.~Huang, Y.~Tang, and R.~Qin, ``An evaluation of planetscope images for 3d reconstruction and change detection--experimental validations with case studies,'' \emph{GIScience \& Remote Sensing}, vol.~59, no.~1, pp. 744--761, 2022.

\bibitem{a6}
J.~Inglada, ``Automatic recognition of man-made objects in high resolution optical remote sensing images by {{SVM}} classification of geometric image features,'' \emph{ISPRS J. Photogramm. Remote Sens.}, vol.~62, no.~3, pp. 236--248, 2007.

\bibitem{a7}
M.~Ortner, X.~Descombes, and J.~Zerubia, ``A {{Marked Point Process}} of {{Rectangles}} and {{Segments}} for {{Automatic Analysis}} of {{Digital Elevation Models}},'' \emph{IEEE Trans. Pattern Anal. Mach. Intell.}, vol.~30, no.~1, pp. 105--119, 2008.

\bibitem{a8}
F.~Lafarge, X.~Descombes, J.~Zerubia, and M.~{Pierrot-Deseilligny}, ``Structural approach for building reconstruction from a single {{DSM}},'' \emph{IEEE Trans. Pattern Anal. Mach. Intell.}, vol.~32, no.~1, pp. 135--147, 2010.

\bibitem{a9}
D.~Cheng, R.~Liao, S.~Fidler, and R.~Urtasun, ``{{DARNet}}: {{Deep Active Ray Network}} for {{Building Segmentation}},'' in \emph{IEEE Conf. Comput. Vis. Pattern Recog.}, 2019, pp. 7423--7431.

\bibitem{a10}
Q.~Zhu, C.~Liao, H.~Hu, X.~Mei, and H.~Li, ``{{MAP-Net}}: {{Multiple Attending Path Neural Network}} for {{Building Footprint Extraction From Remote Sensed Imagery}},'' \emph{{IEEE} Trans. Geosci. Remote Sens.}, vol.~59, no.~7, pp. 6169--6181, 2021.

\bibitem{a11}
M.~Li, F.~Lafarge, and R.~Marlet, ``Approximating shapes in images with low-complexity polygons,'' in \emph{IEEE Conf. Comput. Vis. Pattern Recog.}, 2020, pp. 8630--8638.

\bibitem{a12}
J.~Wang, L.~Meng, W.~Li, W.~Yang, L.~Yu, and G.-S. Xia, ``Learning to {{Extract Building Footprints From Off-Nadir Aerial Images}},'' \emph{IEEE Trans. Pattern Anal. Mach. Intell.}, vol.~45, no.~1, pp. 1294--1301, 2023.

\bibitem{a13}
K.~He, G.~Gkioxari, P.~Dollár, and R.~Girshick, ``Mask {R-CNN},'' in \emph{Int. Conf. Comput. Vis.}, 2017, pp. 2980--2988.

\bibitem{MLSBRN}
W.~Li, H.~Yang, Z.~Hu, J.~Zheng, G.-S. Xia, and C.~He, ``3d building reconstruction from monocular remote sensing images with multi-level supervisions,'' in \emph{IEEE Conf. Comput. Vis. Pattern Recog.}, 2024, pp. 27\,728--27\,737.

\bibitem{a14}
N.~Bodla, B.~Singh, R.~Chellappa, and L.~S. Davis, ``Soft-{NMS} -- {Improving} object detection with one line of code,'' in \emph{Int. Conf. Comput. Vis.}, Oct 2017.

\bibitem{a16}
A.~Kirillov, E.~Mintun, N.~Ravi, H.~Mao, C.~Rolland, L.~Gustafson, T.~Xiao, S.~Whitehead, A.~C. Berg, W.-Y. Lo \emph{et~al.}, ``Segment anything,'' in \emph{Int. Conf. Comput. Vis.}, 2023, pp. 4015--4026.

\bibitem{a17}
L.~Ke, M.~Ye, M.~Danelljan, Y.-W. Tai, C.-K. Tang, F.~Yu \emph{et~al.}, ``Segment anything in high quality,'' \emph{Adv. Neural Inform. Process. Syst.}, vol.~36, 2024.

\bibitem{a18}
K.~Chen, C.~Liu, H.~Chen, H.~Zhang, W.~Li, Z.~Zou, and Z.~Shi, ``Rsprompter: Learning to prompt for remote sensing instance segmentation based on visual foundation model,'' \emph{{IEEE} Trans. Geosci. Remote Sens.}, 2024.

\bibitem{a15}
Y.~He, C.~Zhu, J.~Wang, M.~Savvides, and X.~Zhang, ``Bounding box regression with uncertainty for accurate object detection,'' in \emph{IEEE Conf. Comput. Vis. Pattern Recog.}, 2019.

\bibitem{a19}
G.~Priestnall, J.~Jaafar, and A.~Duncan, ``Extracting urban features from lidar digital surface models,'' \emph{Computers, Environment and Urban Systems}, vol.~24, no.~2, pp. 65--78, 2000.

\bibitem{a20}
S.~R. Khattak, D.~S. Buckstein, and A.~Hogue, ``Reconstructing 3d buildings from lidar using level set methods,'' in \emph{Int. Conf. on Computer and Robot Vision}, 2013, pp. 151--158.

\bibitem{a21}
Z.~Chen, H.~Ledoux, S.~Khademi, and L.~Nan, ``Reconstructing compact building models from point clouds using deep implicit fields,'' \emph{ISPRS J. Photogramm. Remote Sens.}, vol. 194, pp. 58--73, 2022.

\bibitem{b31}
J.~Long, E.~Shelhamer, and T.~Darrell, ``Fully convolutional networks for semantic segmentation,'' in \emph{IEEE Conf. Comput. Vis. Pattern Recog.}, June 2015.

\bibitem{b32}
Z.~Jiang, Y.~Li, C.~Yang, P.~Gao, Y.~Wang, Y.~Tai, and C.~Wang, ``Prototypical contrast adaptation for domain adaptive semantic segmentation,'' in \emph{Eur. Conf. Comput. Vis.}, 2022, pp. 36--54.

\bibitem{b33}
H.~Zhao, J.~Shi, X.~Qi, X.~Wang, and J.~Jia, ``Pyramid scene parsing network,'' in \emph{IEEE Conf. Comput. Vis. Pattern Recog.}, July 2017.

\bibitem{b34}
X.~Zhao, W.~Ding, Y.~An, Y.~Du, T.~Yu, M.~Li, M.~Tang, and J.~Wang, ``Fast segment anything,'' \emph{arXiv preprint arXiv:2306.12156}, 2023.

\bibitem{b35}
Z.~Jiang, Z.~Gu, J.~Peng, H.~Zhou, L.~Liu, Y.~Wang, Y.~Tai, C.~Wang, and L.~Zhang, ``Stc: spatio-temporal contrastive learning for video instance segmentation,'' in \emph{Eur. Conf. Comput. Vis.}\hskip 1em plus 0.5em minus 0.4em\relax Springer, 2022, pp. 539--556.

\bibitem{b36}
X.~Wang, X.~Zhang, Y.~Cao, W.~Wang, C.~Shen, and T.~Huang, ``Seggpt: Segmenting everything in context,'' \emph{arXiv preprint arXiv:2304.03284}, 2023.

\bibitem{b37}
X.~Zou, J.~Yang, H.~Zhang, F.~Li, L.~Li, J.~Wang, L.~Wang, J.~Gao, and Y.~J. Lee, ``Segment everything everywhere all at once,'' \emph{Adv. Neural Inform. Process. Syst.}, vol.~36, 2024.

\bibitem{b29}
R.~Zhang, Z.~Jiang, Z.~Guo, S.~Yan, J.~Pan, X.~Ma, H.~Dong, P.~Gao, and H.~Li, ``Personalize {{Segment Anything Model}} with {{One Shot}},'' Oct. 2023.

\bibitem{b30}
N.~Ruiz, Y.~Li, V.~Jampani, Y.~Pritch, M.~Rubinstein, and K.~Aberman, ``Dreambooth: Fine tuning text-to-image diffusion models for subject-driven generation,'' in \emph{IEEE Conf. Comput. Vis. Pattern Recog.}, 2023, pp. 22\,500--22\,510.

\bibitem{b38}
F.~Raji{\v{c}}, L.~Ke, Y.-W. Tai, C.-K. Tang, M.~Danelljan, and F.~Yu, ``Segment anything meets point tracking,'' \emph{arXiv preprint arXiv:2307.01197}, 2023.

\bibitem{b39}
N.~Karaev, I.~Rocco, B.~Graham, N.~Neverova, A.~Vedaldi, and C.~Rupprecht, ``Cotracker: It is better to track together,'' \emph{arXiv preprint arXiv:2307.07635}, 2023.

\bibitem{b40}
T.~Chen, L.~Zhu, C.~Deng, R.~Cao, Y.~Wang, S.~Zhang, Z.~Li, L.~Sun, Y.~Zang, and P.~Mao, ``Sam-adapter: Adapting segment anything in underperformed scenes,'' in \emph{Int. Conf. Comput. Vis.}, October 2023, pp. 3367--3375.

\bibitem{b28}
L.~P. Osco, Q.~Wu, E.~L. {de Lemos}, W.~N. Gon{\c c}alves, A.~P.~M. Ramos, J.~Li, and J.~Marcato, ``The {{Segment Anything Model}} ({{SAM}}) for remote sensing applications: {{From}} zero to one shot,'' \emph{Int. J. of Applied Earth Observation and Geoinformation}, vol. 124, p. 103540, 2023.

\bibitem{yu2023h2rboxv2}
Y.~Yu, X.~Yang, Q.~Li, Y.~Zhou, G.~Zhang, J.~Yan, and F.~Da, ``H2rbox-v2: Boosting hbox-supervised oriented object detection via symmetric learning,'' \emph{arXiv preprint arXiv:2304.04403}, 2023.

\bibitem{b41}
P.~Liu, X.~Liu, M.~Liu, Q.~Shi, J.~Yang, X.~Xu, and Y.~Zhang, ``Building footprint extraction from high-resolution images via spatial residual inception convolutional neural network,'' \emph{Remote Sens.}, vol.~11, no.~7, p. 830, 2019.

\bibitem{b42}
J.~Kang, R.~Fernandez-Beltran, X.~Sun, J.~Ni, and A.~Plaza, ``Deep learning-based building footprint extraction with missing annotations,'' \emph{{IEEE} Geosci. Remote Sens. Lett.}, vol.~19, pp. 1--5, 2021.

\bibitem{b43}
W.~Li, C.~He, J.~Fang, J.~Zheng, H.~Fu, and L.~Yu, ``Semantic segmentation-based building footprint extraction using very high-resolution satellite images and multi-source gis data,'' \emph{Remote Sens.}, vol.~11, no.~4, p. 403, 2019.

\bibitem{b44}
T.~Yu, P.~Tang, B.~Zhao, S.~Bai, P.~Gou, J.~Liao, and C.~Jin, ``Convbnet: A convolutional network for building footprint extraction,'' \emph{{IEEE} Geosci. Remote Sens. Lett.}, vol.~20, pp. 1--5, 2023.

\bibitem{b45}
J.~Cai and Y.~Chen, ``Mha-net: Multipath hybrid attention network for building footprint extraction from high-resolution remote sensing imagery,'' \emph{{IEEE} J. Sel. Topics Appl. Earth Observations Remote Sens.}, vol.~14, pp. 5807--5817, 2021.

\bibitem{darnet}
D.~Cheng, R.~Liao, S.~Fidler, and R.~Urtasun, ``Darnet: Deep active ray network for building segmentation,'' in \emph{IEEE Conf. Comput. Vis. Pattern Recog.}, June 2019.

\bibitem{ma2023local}
Z.~Ma, M.~Xia, L.~Weng, and H.~Lin, ``Local feature search network for building and water segmentation of remote sensing image,'' \emph{Sustainability}, vol.~15, no.~4, p. 3034, 2023.

\bibitem{shi2020building}
Y.~Shi, Q.~Li, and X.~X. Zhu, ``Building segmentation through a gated graph convolutional neural network with deep structured feature embedding,'' \emph{ISPRS J. Photogramm. Remote Sens.}, vol. 159, pp. 184--197, 2020.

\bibitem{sariturk2022residual}
B.~Sariturk and D.~Z. Seker, ``A residual-inception u-net (riu-net) approach and comparisons with u-shaped cnn and transformer models for building segmentation from high-resolution satellite images,'' \emph{Sensors}, vol.~22, no.~19, p. 7624, 2022.

\bibitem{christie2020unet}
G.~Christie, R.~R. R.~M. Abujder, K.~Foster, S.~Hagstrom, G.~D. Hager, and M.~Z. Brown, ``Learning geocentric object pose in oblique monocular images,'' in \emph{IEEE Conf. Comput. Vis. Pattern Recog.}, 2020, pp. 14\,512--14\,520.

\bibitem{MTBRNet}
W.~Li, L.~Meng, J.~Wang, C.~He, G.-S. Xia, and D.~Lin, ``3d building reconstruction from monocular remote sensing images,'' in \emph{Int. Conf. Comput. Vis.}, 2021, pp. 12\,548--12\,557.

\bibitem{a27}
S.~Liu, F.~Li, H.~Zhang, X.~Yang, X.~Qi, H.~Su, J.~Zhu, and L.~Zhang, ``Dab-detr: Dynamic anchor boxes are better queries for detr,'' in \emph{Int. Conf. Learn. Represent.}, 2022.

\bibitem{a23}
N.~Carion, F.~Massa, G.~Synnaeve, N.~Usunier, A.~Kirillov, and S.~Zagoruyko, ``End-to-end object detection with transformers,'' in \emph{Eur. Conf. Comput. Vis.}, 2020, pp. 213--229.

\bibitem{maskformer}
B.~Cheng, A.~Schwing, and A.~Kirillov, ``Per-pixel classification is not all you need for semantic segmentation,'' vol.~34, pp. 17\,864--17\,875, 2021.

\bibitem{masktwoformer}
B.~Cheng, I.~Misra, A.~G. Schwing, A.~Kirillov, and R.~Girdhar, ``Masked-attention mask transformer for universal image segmentation,'' in \emph{IEEE Conf. Comput. Vis. Pattern Recog.}, 2022.

\bibitem{dn}
F.~Li, H.~Zhang, S.~Liu, J.~Guo, L.~M. Ni, and L.~Zhang, ``Dn-detr: Accelerate detr training by introducing query denoising,'' in \emph{IEEE Conf. Comput. Vis. Pattern Recog.}, 2022, pp. 13\,619--13\,627.

\bibitem{dino}
H.~Zhang, F.~Li, S.~Liu, L.~Zhang, H.~Su, J.~Zhu, L.~M. Ni, and H.-Y. Shum, ``Dino: Detr with improved denoising anchor boxes for end-to-end object detection,'' 2022.

\bibitem{maskdino}
F.~Li, H.~Zhang, H.~Xu, S.~Liu, L.~Zhang, L.~M. Ni, and H.-Y. Shum, ``Mask dino: Towards a unified transformer-based framework for object detection and segmentation,'' in \emph{IEEE Conf. Comput. Vis. Pattern Recog.}, 2023, pp. 3041--3050.

\bibitem{zhang2021survey}
Y.~Zhang and Q.~Yang, ``A survey on multi-task learning,'' \emph{{IEEE} Trans. Knowl. Data Eng.}, vol.~34, no.~12, pp. 5586--5609, 2021.

\bibitem{htc}
K.~Chen, J.~Pang, J.~Wang, Y.~Xiong, X.~Li, S.~Sun, W.~Feng, Z.~Liu, J.~Shi, W.~Ouyang, C.~C. Loy, and D.~Lin, ``Hybrid task cascade for instance segmentation,'' in \emph{IEEE Conf. Comput. Vis. Pattern Recog.}, 2019.

\bibitem{li2020automatic}
P.~Li, Y.~Li, J.~Feng, Z.~Ma, and X.~Li, ``Automatic detection and recognition of road intersections for road extraction from imagery,'' \emph{Int. Arch. Photogramm. Remote Sens. Spatial Inf. Sci.}, vol.~43, pp. 113--117, 2020.

\bibitem{eem}
Y.~Liu, Z.~Han, C.~Chen, L.~Ding, and Y.~Liu, ``Eagle-eyed multitask cnns for aerial image retrieval and scene classification,'' \emph{{IEEE} Trans. Geosci. Remote Sens.}, vol.~58, no.~9, pp. 6699--6721, 2020.

\bibitem{drive}
\BIBentryALTinterwordspacing
W.~Jun, M.~Son, J.~Yoo, and S.~Lee, ``Optimal configuration of multi-task learning for autonomous driving,'' \emph{Sensors}, vol.~23, no.~24, 2023. [Online]. Available: \url{https://www.mdpi.com/1424-8220/23/24/9729}
\BIBentrySTDinterwordspacing

\bibitem{liu2021abnet}
Y.~Liu, Q.~Li, Y.~Yuan, Q.~Du, and Q.~Wang, ``Abnet: Adaptive balanced network for multiscale object detection in remote sensing imagery,'' \emph{{IEEE} Trans. Geosci. Remote Sens.}, vol.~60, pp. 1--14, 2022.

\bibitem{hybridnet}
D.~Vu, B.~Ngo, and H.~Phan, ``Hybridnets: End-to-end perception network,'' \emph{arXiv preprint arXiv:2203.09035}, 2022.

\bibitem{yolop}
D.~Wu, M.-W. Liao, W.-T. Zhang, X.-G. Wang, X.~Bai, W.-Q. Cheng, and W.-Y. Liu, ``Yolop: You only look once for panoptic driving perception,'' \emph{Mach. Intell. Research}, vol.~19, no.~6, pp. 550--562, 2022.

\bibitem{yolopv2}
C.~Han, Q.~Zhao, S.~Zhang, Y.~Chen, Z.~Zhang, and J.~Yuan, ``Yolopv2: Better, faster, stronger for panoptic driving perception,'' \emph{arXiv preprint arXiv:2208.11434}, 2022.

\bibitem{wang2022hybrid}
Q.~Wang, Y.~Liu, Z.~Xiong, and Y.~Yuan, ``Hybrid feature aligned network for salient object detection in optical remote sensing imagery,'' \emph{{IEEE} Trans. Geosci. Remote Sens.}, vol.~60, pp. 1--15, 2022.

\bibitem{SDNet}
Y.~Liu, Z.~Xiong, Y.~Yuan, and Q.~Wang, ``Transcending pixels: Boosting saliency detection via scene understanding from aerial imagery,'' \emph{{IEEE} Trans. Geosci. Remote Sens.}, vol.~61, pp. 1--16, 2023.

\bibitem{a22}
K.~He, X.~Chen, S.~Xie, Y.~Li, P.~Doll\'ar, and R.~Girshick, ``Masked autoencoders are scalable vision learners,'' in \emph{IEEE Conf. Comput. Vis. Pattern Recog.}, 2022, pp. 16\,000--16\,009.

\bibitem{a24}
G.~Cheng, X.~Yuan, X.~Yao, K.~Yan, Q.~Zeng, X.~Xie, and J.~Han, ``Towards large-scale small object detection: Survey and benchmarks,'' \emph{IEEE Trans. Pattern Anal. Mach. Intell.}, 2023.

\bibitem{celoss}
C.~E. Shannon, ``A mathematical theory of communication,'' \emph{Bell Syst. Tech. J.}, vol.~27, no.~3, pp. 379--423, 1948.

\bibitem{fastrcnn}
R.~Girshick, ``Fast r-cnn,'' in \emph{Int. Conf. Comput. Vis.}, 2015, pp. 1440--1448.

\bibitem{a25}
T.-Y. Lin, M.~Maire, S.~Belongie, J.~Hays, P.~Perona, D.~Ramanan, P.~Doll{\'a}r, and C.~L. Zitnick, ``Microsoft coco: Common objects in context,'' in \emph{Eur. Conf. Comput. Vis.}, 2014, pp. 740--755.

\bibitem{a26}
B.~Cheng, R.~Girshick, P.~Doll{\'a}r, A.~C. Berg, and A.~Kirillov, ``Boundary iou: Improving object-centric image segmentation evaluation,'' in \emph{IEEE Conf. Comput. Vis. Pattern Recog.}, 2021, pp. 15\,334--15\,342.

\bibitem{mmdetection}
K.~Chen, J.~Wang, J.~Pang, Y.~Cao, Y.~Xiong, X.~Li, S.~Sun, W.~Feng, Z.~Liu, J.~Xu, Z.~Zhang, D.~Cheng, C.~Zhu, T.~Cheng, Q.~Zhao, B.~Li, X.~Lu, R.~Zhu, Y.~Wu, J.~Dai, J.~Wang, J.~Shi, W.~Ouyang, C.~C. Loy, and D.~Lin, ``{MMDetection}: Open mmlab detection toolbox and benchmark,'' \emph{arXiv preprint arXiv:1906.07155}, 2019.

\bibitem{sgd}
H.~Robbins and S.~Monro, ``A stochastic approximation method,'' \emph{Ann. Math. Stat.}, pp. 400--407, 1951.

\bibitem{gradclip}
R.~Pascanu, T.~Mikolov, and Y.~Bengio, ``On the difficulty of training recurrent neural networks,'' in \emph{Int. Conf. on Mach. Learn.}\hskip 1em plus 0.5em minus 0.4em\relax Pmlr, 2013, pp. 1310--1318.

\bibitem{omnicity}
W.~Li, Y.~Lai, L.~Xu, Y.~Xiangli, J.~Yu, C.~He, G.-S. Xia, and D.~Lin, ``Omnicity: Omnipotent city understanding with multi-level and multi-view images,'' in \emph{IEEE Conf. Comput. Vis. Pattern Recog.}, 2023, pp. 17\,397--17\,407.

\end{thebibliography}

\end{document}